\documentclass[10pt,twocolumn,letterpaper]{article}
\usepackage{cvpr, times}
\PassOptionsToPackage{numbers, compress}{natbib}
\cvprfinalcopy

\usepackage[ruled,linesnumbered]{algorithm2e}
\usepackage{xcolor}
\usepackage[utf8]{inputenc} 
\usepackage[T1]{fontenc}    
\usepackage[pagebackref=true,breaklinks=true,letterpaper=true,colorlinks,bookmarks=false]{hyperref}
\usepackage{url}            
\usepackage{booktabs}       
\usepackage{amsfonts}       
\usepackage{nicefrac}       
\usepackage{microtype}      
\usepackage{color}          
\usepackage{enumitem}
\usepackage{amsmath}
\usepackage{bbold}
\usepackage{diagbox}
\usepackage{graphbox}
\usepackage{float}
\usepackage{listings}
\usepackage{graphicx}
\usepackage[caption=false]{subfig}
\usepackage{sidecap}
\usepackage{setspace}
\usepackage{tabularx}
\usepackage{wrapfig}
\usepackage{amssymb}
\usepackage{multirow}
\usepackage{wrapfig,lipsum,booktabs}
\newtheorem{theorem}{Theorem}  

\def\cvprPaperID{3255}
\title{Binary Ensemble Neural Network:\\ More Bits per Network or More Networks per Bit?}

\begin{document}
\author{
  Shilin Zhu \\
  UC San Diego\\
  La Jolla, CA 92093 \\
  {\tt\small shz338@eng.ucsd.edu} \\
  \and
  Xin Dong \\
  Harvard University\\
  Cambridge, MA 02138 \\
  {\tt\small xindong@g.harvard.edu} \\
  \and
  Hao Su \\
  UC San Diego\\
  La Jolla, CA 92093 \\
  {\tt\small haosu@eng.ucsd.edu} \\
}

\maketitle

\begin{abstract}
    Binary neural networks (BNN) have been studied extensively since they run dramatically faster at lower memory and power consumption than floating-point networks, thanks to the efficiency of bit operations. However, contemporary BNNs whose weights and activations are both single bits suffer from severe accuracy degradation. To understand why, we investigate the representation ability, speed and bias/variance of BNNs through extensive experiments. We conclude that the error of BNNs are predominantly caused by the intrinsic instability (training time) and non-robustness (train \& test time). Inspired by this investigation, we propose the Binary Ensemble Neural Network (BENN) which leverages ensemble methods to improve the performance of BNNs with limited efficiency cost. While ensemble techniques have been broadly believed to be only marginally helpful for strong classifiers such as deep neural networks, our analysis and experiments show that they are naturally a perfect fit to boost BNNs. We find that our BENN, which is faster and more robust than state-of-the-art binary networks, can even surpass the accuracy of the full-precision floating number network with the same architecture. 
\end{abstract}
\vspace{-6mm}
\section{Introduction}
\vspace{-2mm}

Deep Neural Networks (DNNs) have achieved great impact to broad disciplines in academia and industry~\cite{szegedy2015going, krizhevsky2012imagenet}. Recently, the deployment of DNNs are transferring from high-end cloud to low-end devices such as mobile phones and embedded chips, serving general public with many real-time applications, such as drones, miniature robots, and augmented reality.  
Unfortunately, these devices typically have limited computing power and memory space, thus cannot afford DNNs to achieve important tasks like object recognition involving significant matrix computation and memory usage. 

Binary Neural Network (BNN) is among the most promising techniques to meet the desired computation and memory requirement. BNNs~\cite{hubara2016binarized} are deep neural networks whose weights and activations have only two possible values (e.g., -1 and +1) and can be represented by a single bit. Beyond the obvious advantage of saving storage and memory space, the binarized architecture admits only bitwise operations, which can be computed extremely fast using digital logic units~\cite{govindu2004analysis} such as arithmetic-logic unit (ALU) with much less power consumption than floating-point unit (FPU). 

\begin{figure}
    \vspace{2mm}
    \centering
    \includegraphics[width=\linewidth]{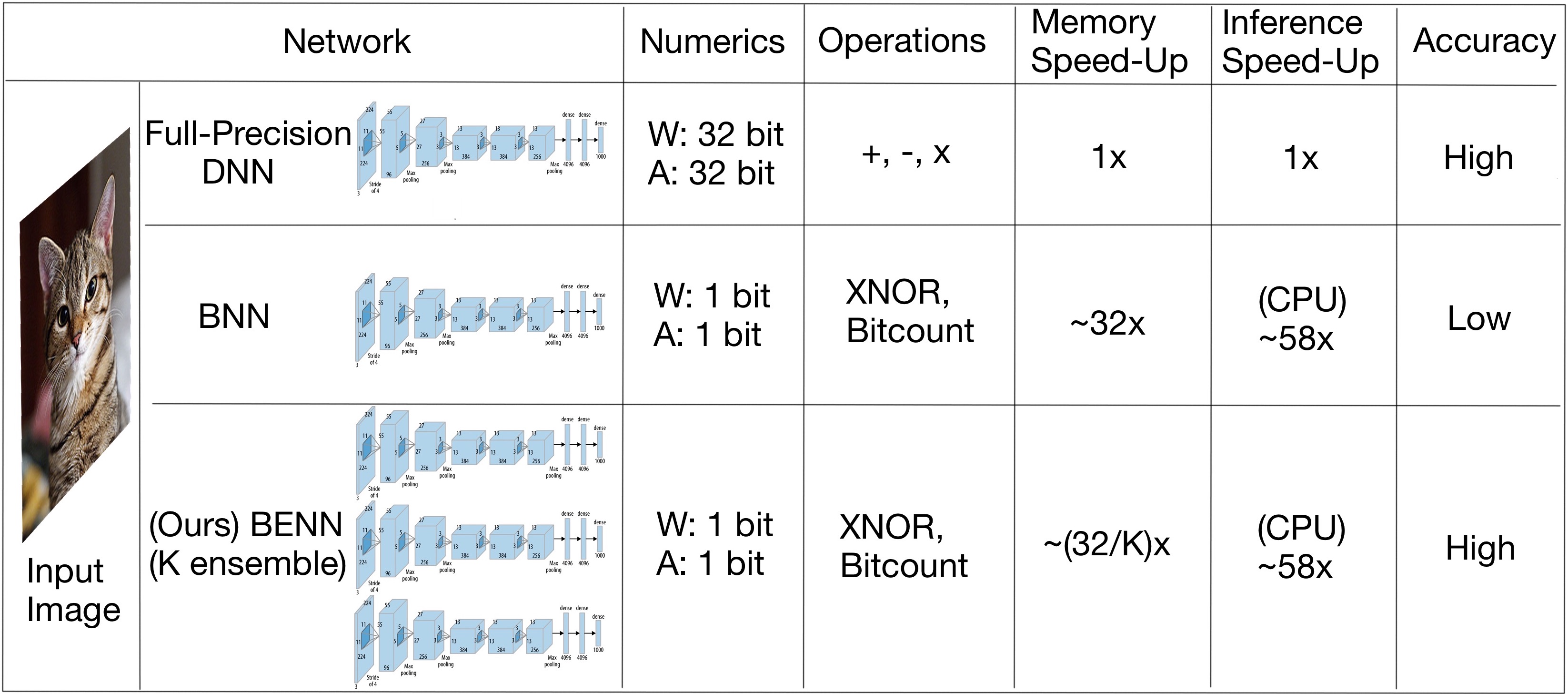}
    \vspace{-3mm}
    \caption{Comparison between traditional floating-number DNN, BNN and our proposed BENN on image recognition task (W: weights, A: activations). The inference speed of BENN can be further boosted on FPGAs \cite{zhao2017accelerating}.}
    \label{fig:benn}
    \vspace{-5mm}
\end{figure}

Despite the significant gain in speed and storage, however, current BNNs suffer from notable accuracy degradation when applied to challenging tasks such as ImageNet classification. To mitigate the gap, previous researches in BNNs have been focusing on designing more effective optimization algorithms to find better local minima of the quantized weights. However, the task is highly non-trivial, since gradient-based optimization that used to be effective to train DNNs now becomes tricky to implement. 

Instead of tweaking network optimizers, we investigate BNNs systematically in terms of representation power, speed, bias, variance, stability, and their robustness. We find that BNNs suffer from severe intrinsic instability and non-robustness regardless of network parameter values. What implied by this observation is that the performance degradation of BNNs are not likely to be resolved by solely improving the optimization techniques; instead, it is mandatory to cure the BNN function, particularly to reduce the prediction variance and improve its robustness to noises.

Inspired by the analysis, in this work, we propose Binary Ensemble Neural Network (BENN). Though the basic idea is as straight-forward as to simply aggregate multiple BNNs by boosting or bagging, we show that the statistical properties of the ensembled classifiers become much nicer: not only the bias and variance are reduced, more importantly, BENN's robustness to noises at test time is significantly improved. All the experiments suggest that BNNs and ensemble methods are a perfectly natural fit. Using architectures of the same connectivity (a compact Network in Network \cite{lin2013network}), we find that \emph{boosting only $4 \sim 5$ BNNs would be able to even surpass the baseline DNN with real weights in the best case.} In addition, our initial exploration by applying BENN on ImageNet recognition using AlexNet \cite{krizhevsky2012imagenet} and ResNet \cite{he2016deep} also shows a large gain. This is by far the fastest, most accurate, and most robust results achieved by binarized networks (Fig.~\ref{fig:benn}).

To the best of our knowledge, this is the first work to bridge BNNs with ensemble methods. Unlike traditional BNN improvements that have computational complexity of $\gtrsim$ $O(K^{2})$ by using $K$-bit per weights~\cite{zhou2016dorefa} or $K$ bases in total~\cite{lin2017towards}, the complexity of BENN is reduced to $O(K)$. Compared with \cite{zhou2016dorefa, lin2017towards}, BENN also enjoys better bitwise operation parallelizability. With trivial parallelization, the complexity can be reduced to $O(1)$.  We believe that BENN can shed light on more research along this idea to achieve extremely fast yet robust computation by networks. 

\section{Related Work}

\textbf{Quantized and binary neural networks: } People have found that there is no need to use full-precision parameters and activations and can still preserve the accuracy of a neural network using k-bit fixed point numbers, as stated by \cite{gong2014compressing, han2015deep, wu2016quantized, cai2017deep, li2017training, lin2016fixed, park2017weighted, sung2015resiliency, polino2018model}. The first approach is to use low-bitwidth numbers to approximate real ones, which is called quantized neural networks (QNNs) \cite{hubara2016quantized}. \cite{zhu2016trained, zhou2017incremental} also proposed ternary neural networks. Although recent advances such as \cite{zhou2016dorefa} can achieve competitive performance compared with full-precision models, they cannot fully speed it up because we still cannot perform parallelized bitwise operation with bitwidth larger than one. \cite{hubara2016binarized} is the very recent work that binarizes all the weights and activations, which was the birth of BNN. They have demonstrated the power of BNNs in terms of speed, memory use and power consumption. But recent works such as \cite{tang2017train, courbariaux2016binarized, guo2017network, courbariaux2015binaryconnect} also reveal the strong accuracy degradation and mismatch issue during the training when BNNs are applied in complicated tasks such as ImageNet (\cite{deng2009imagenet}) recognition, especially when the activation is binarized. Although some work like \cite{lin2017towards, rastegari2016xnor, deng2017gated} have offered reasonable solutions to approximate full-precision neural network, much more computation and tricks on hyperparameters are still needed to implement compared with BENN. Since they either use $K$-bitwidth quantization or $K$ binary bases, the computational complexity cannot get rid of $O(K^{2})$ if $O(1)$ is required for 1-bit single BNN, while BENN can achieve $O(K)$ and even $O(1)$ if multiple threads are naturally paralleled. Also, many of current literatures tried to minimize the distance between binary and real-value parameters. But empirical assumptions such as Gaussian parameter distribution are usually required in order to get a priori for each BNN or just keep the sign same as suggested by \cite{lin2017towards}, otherwise the non-convex optimization is hard to deal with. By contrast, BENN can be a general framework to achieve the goal and has strong potential to work even better than full-precision networks, without involving more hyperparameters than a single BNN.

\textbf{Ensemble techniques: }To avoid simply relying on a single powerful classifier, the ensemble strategy can improve the accuracy of given learning algorithm combining multiple weak classifiers as summarized by \cite{breiman1996bias, carney1999confidence, oza2001online}. The two most common strategies are bagging by \cite{breiman1996bagging} and boosting by \cite{schapire2003boosting, friedman2000additive, schapire1999improved, hastie2009multi}, which were proposed many years ago and have strong statistical foundation. They have roots in a theoretical framework PAC model by \cite{valiant1984theory} which was the first to pose the question of whether weak learners can be ensembled into a strong learner. Bagging predictors are proved to reduce variance while boosting can reduce both bias and variance, and their effectiveness have been proved by many theoretical analysis. Traditionally ensemble was used with decision trees, decision stumps, random forests and achieved great success thanks to its desirable statistical properties. Recently people use ensemble to increase the generalization ability of deep CNNs \cite{han2016incremental}, advocate boosting on CNNs and do architecture selection \cite{moghimi2016boosted}, and propose boost over features \cite{huang2017learning}. But people did not pay enough attention to ensemble techniques because neural network is not a weak classifier anymore thus ensemble can unnecessarily increase the model complexity. However, when applied to weak binary neural networks, we found it generates new insights and hopes, and BENN is a natural outcome of such perfect combination. In this work, we build our BENN on the top of variant bagging, AdaBoost by  \cite{freund1995desicion, schapire2013explaining}, LogitBoost by \cite{friedman2000additive} and can be extended to many more variants of traditional ensemble algorithms. We hope this work can revive these intelligent approaches and bring their life back into modern neural networks. 

\section{Why Making BNNs Work Well is Challenging?}
\label{sec:inspiration}
Despite the speed and space advantage of BNN, its performances is still far inferior to the real valued counterparts. There are at least two possible reasons: First, functions representable by BNNs may have some inherent flaws; Second, current optimization algorithms may still not be able to find a good minima. While most researchers have been working on developing better optimization methods, we suspect that BNNs have some fundamental flaws. The following investigation reveals the fundamental limitations of BNN-representable functions experimentally. 

Because all weights and activations are binary, an obvious fact is that BNNs can only represent a subset of discrete functions, being strictly weaker than real networks that are universal continuous function approximators~\cite{hornik1989multilayer}. What are not so obvious are two serious limitations of BNNs: the robustness issue w.r.t. input perturbations, and the stability issue w.r.t. network parameters. Classical learning theory tells us that both robustness and stability are closely related to the generalization error of a model \cite{xu2012robustness, bousquet2002stability}. A more detailed theoretical analysis on BNN's problems is attached in supplementary material.

\textbf{Robustness Issue:} In practice, we observe more severe overfitting effects of BNNs than real networks. 
Robustness is defined as the property that if a testing population is ``similar'' to a training population, then the testing error is close to the training error \cite{xu2012robustness}. To verify this point, we experiment in a random network setting and a trained network setting.

\emph{Random Network Setting.} We compute the following quantity to compare 32bit real-valued DNN, BNN, QNN, and our BENN model (Sec.~\ref{sec:method}) on the Network-In-Network (NIN) architecture:
\vspace{-1mm}
\begin{equation} \label{eq:1}
    \mathbb E_{w}\mathbb E_{\Delta x}||f(x+\Delta x;w)-f(x;w)||^{2}
    \vspace{-1mm}
\end{equation}
where $f$ is the network and $w$ represents network weights. 

We randomly sample real-valued weights $w\sim \mathcal{N}(\mathbb{0}, I)$ as suggested in literature to get a DNN $f_r$ with weights $w_r$ and binarize it to get a BNN $f_b$ with binary weights $w_{b}$. We also independently sample and binarize $w_r$ to generate multiple BNNs with the same architecture to simulate the BENN and get $w_{benn}$. QNN is obtained by quantizing the DNN to $k$-bit weights (W) and activations (A). We normalize each input image in CIFAR-10 to the range $[-1,1]$. 

\begin{figure*}[ht!]
    \vspace{-2mm}
    \minipage{0.33\textwidth}
    \includegraphics[width=\linewidth]{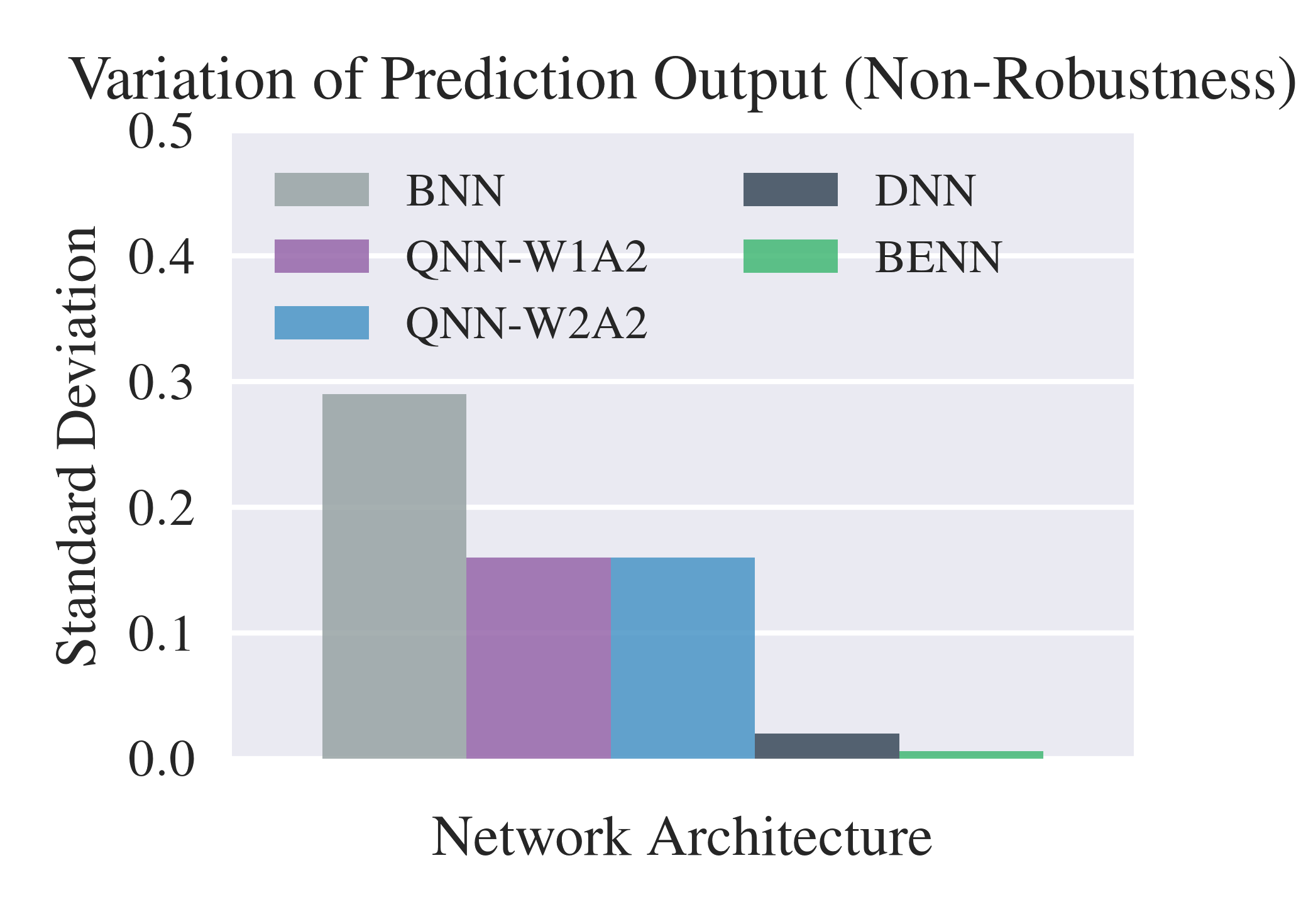}
    \label{fig:pert_1}
    \endminipage\hfill
    \minipage{0.33\textwidth}
    \includegraphics[width=\linewidth]{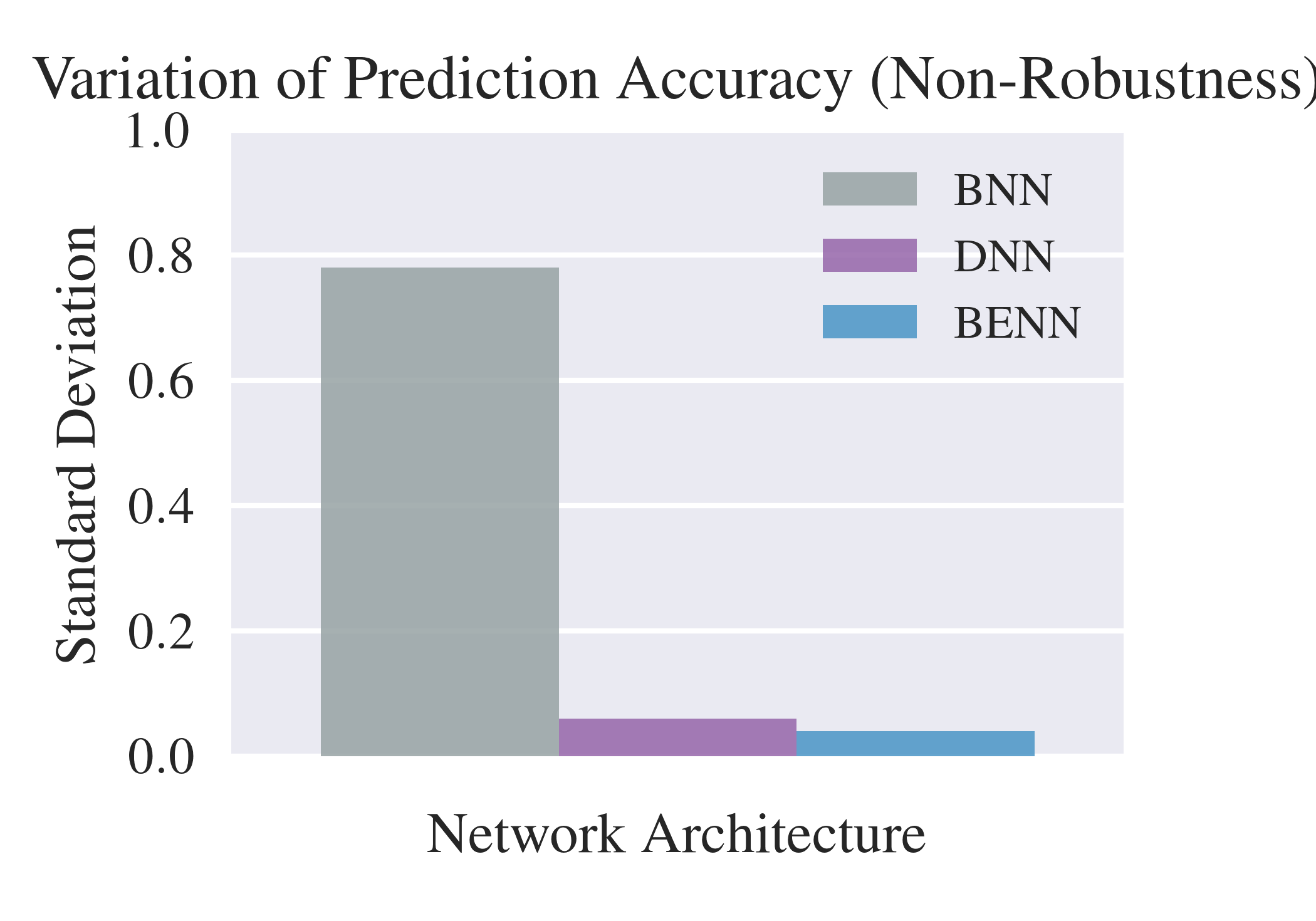}
    \label{fig:pert_2}
    \endminipage\hfill
    \minipage{0.33\textwidth}
    \includegraphics[width=\linewidth]{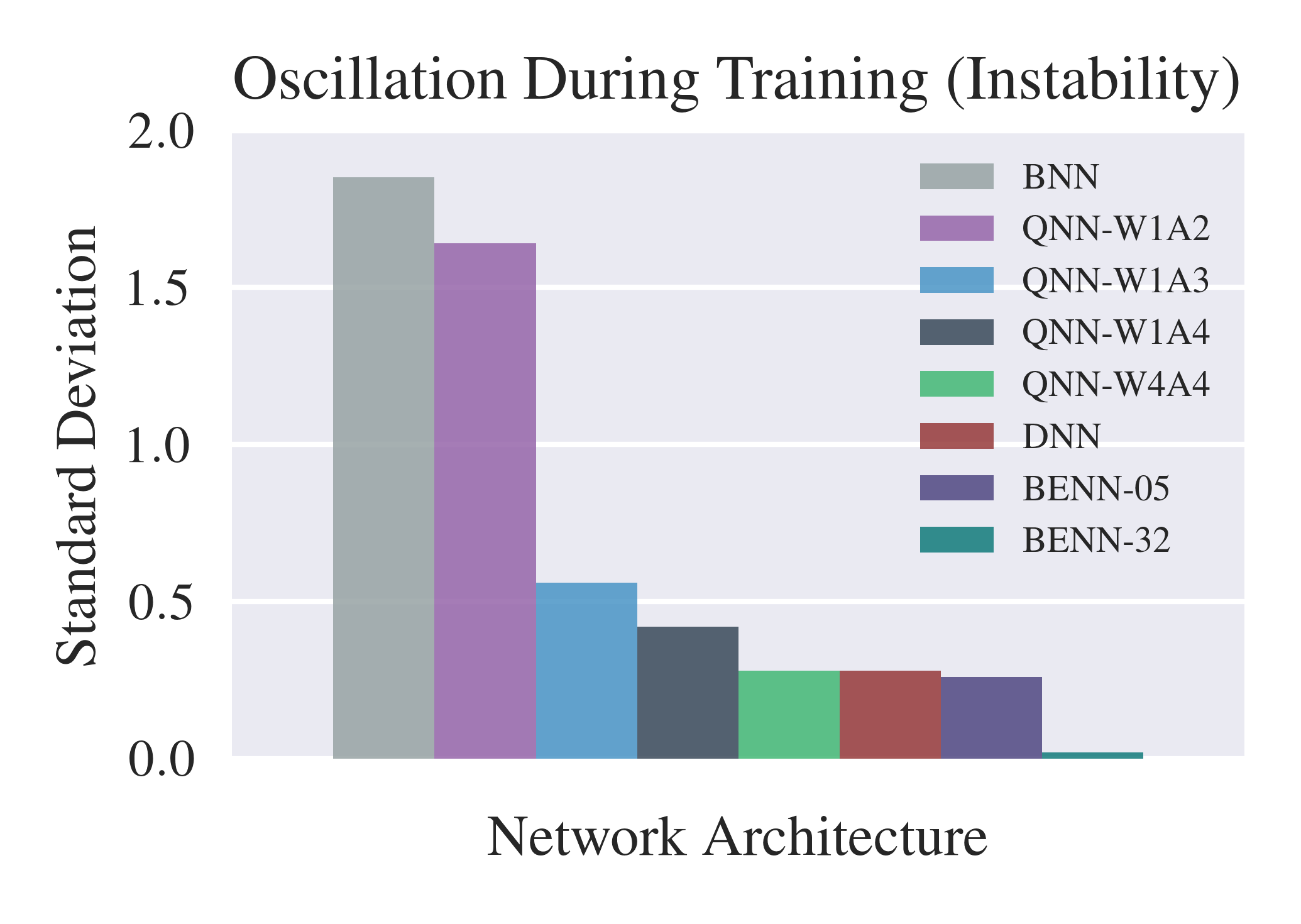}
    \label{fig:pert_2}
    \endminipage\hfill
    \vspace{-3mm}
    \caption{{\small \textbf{Left: }BNN has large output variation (robustness issue). \textbf{Middle: }BNN has large variation of prediction accuracy (robustness issue). \textbf{Right: }BNN has large test accuracy variation during training (instability issue). BENN can cure these problems. Here, the perturbation variance is 0.01. \textit{(*QNN-W1A2 denotes QNN with 1-bit weights and 2-bit activations and so do others.)}}}
    \label{fig:BNN_pert}
    \vspace{-5mm}
\end{figure*}

Then we inject the input perturbation $\Delta x$ on each example by a Gaussian noise with different variances ($0.001 \sim 0.1$), run a forward pass on each network, and measure the expected $l_{2}$ norm of the change on the output distribution.
The above $l_{2}$ norm of DNN, BNN, QNN, and BENN averaged by 1000 sampling rounds is shown in Fig.~\ref{fig:BNN_pert}(left) with perturbation variance 0.01. 

Results show that BNNs always have larger output variation, suggesting that they are more susceptible to input perturbation, and BNN does worse than QNN that has more bits. We also observe that having more bits on activations actually improves BNN's robustness significantly, while having more bits on weights just has quite marginal improvement (Fig.~\ref{fig:BNN_pert}(left)). Therefore, the activation binarization seems to be the bottleneck. 

\emph{Trained Network Setting.} To further consolidate the discovery, we also train a real-valued DNN $f_{r}$ and a BNN $f_{b}$ using XNOR-Net \cite{rastegari2016xnor} rather than direct sampling. We also include our designed BENN $f_{benn}$ in comparison. Then we perform the same Gaussian input perturbation $\Delta x$, run a forward pass, and calculate the change of classification error $\mathcal{L}$ on CIFAR-10 as:
\vspace{-1mm}
\begin{equation}
    \mathbb E_{\Delta x}||\mathcal{L}(f(x+\Delta x))-\mathcal{L}(f(x))||^{2}
    \vspace{-1mm}
\end{equation}
Results in Fig.~\ref{fig:BNN_pert}(middle) indicates that BNNs are still more sensitive to noises even if it is well optimized. 
Although people have shown that weights in BNN still have nice statistical properties as in \cite{anderson2017high}, the conclusion can change dramatically if both weights and activations are binarized while input is perturbed.

\textbf{Stability Issue:} BNNs are known to be hard to optimize due to problems such as gradient mismatch and non-smoothness of activation function. While \cite{li2017training} has shown that stochastic rounding converges to within $O(\Delta)$ accuracy of the minimizer in expectation where $\Delta$ denotes quantization resolution, assuming the error surface is convex,
the community has not fully understood the non-convex error surface of BNN and how it interacts with different optimizers such as SGD or ADAM \cite{kingma2014adam}. 

To compare the stability of different networks (sensitivity to network parameter during optimization), we measure the accuracy fluctuation after a large amount of training steps. Fig.~\ref{fig:BNN_pert} (right) shows the accuracy oscillation in the last 20 training steps after we train BNN and QNN with 300 epochs, and results show that we should at least have QNN with weights and activations both 4-bit in order to stabilize the network.

One explanation of such instability is the non-smoothness of the function output w.r.t. the binary network parameters. Note that, as the output of the activation function in the previous layer, the input to each layer of BNNs are binarized numbers. In other words, not only each function is non-smooth w.r.t. the input, but also it is non-smooth w.r.t. the learned parameters.  As a comparison, empirically, BENN with 5 and 32 ensembles (denoted as BENN-05/32 in Fig.~\ref{fig:BNN_pert}) have already achieved amazing stability.


\section{Binary Ensemble Neural Network}

\label{sec:method}
In this section, we illustrate our BENN using bagging and boosting strategies, respectively. 
In all experiments, we adopt the widely used \emph{deterministic binarization} as $x_{b} = \text{Sign}(x)$ for network weights and activations, which is preferred to leverage hardware accelerations. 
However, back-propagation becomes challenging since the derivative is zero almost everywhere except for the stepping point. In this work, we borrow the common strategy called ``straight-through estimator'' (STE)~\cite{hinton} during back-propagation, defined as $\frac{\partial J}{\partial x} = \frac{\partial J}{\partial x_{b}}I_{|x| \leq 1}$. 

\subsection{BENN-Bagging}

The key idea of bagging is to average weak classifiers that are trained from i.i.d. samples of the training set.
To train each BNN classifier, we sample $M$ examples independently with replacement from the training set $\mathcal{D}$. We do this $K$ times to get $K$ BNNs, denoted as $b^{1}, ..., b^{K}$. The sampling with replacement assures that each BNN sees roughly $63\%$ of the entire training set.

At test time, we aggregate the opinions from these $K$ classifiers and decide among $C$ classes. We compare two ways of aggregating the outputs. One is to choose the label that most BNNs agree with (hard decision), while the other is to choose the best label after aggregating their softmax probabilities (soft decision). 

The main advantage brought by bagging is to reduce the variance of a single classifier. This is known to be extremely effective for deep decision trees which suffer from high variance, but only marginally helpful to boost the performance of neural networks, since networks are generally quite stable. 
Interestingly, though less helpful to real-valued networks, bagging is effective to improve BNNs since the instability issue is severe for BNNs due to gradient mismatch and strong discretization noise as stated in Sec.~\ref{sec:inspiration}. 


\subsection{BENN-Boosting}  

Boosting is another important tool to ensemble classifiers. Instead of just aggregating the predictions from multiple independently trained BNNs, boosting combines multiple weak classifiers in a sequential manner and can be viewed as a stage-wise gradient descent method optimized in the function space. Boosting is able to reduce both bias and variance of individual classifiers.  

There are many variants of boosting algorithms and we choose the AdaBoost~\cite{freund1995desicion} algorithm for its popularity. Suppose classifier $k$ has hypothesis $b^{k}:X \rightarrow \mathbb R$, weight $\alpha_{k}$, and output distribution $p^{k}$, we can denote the aggregated classifier as $B^{K}:X \rightarrow \mathbb R$ and its aggregated output distribution $P^{K}$. Then AdaBoost minimizes the following exponential loss:
\[
    J(B^{K}) = \sum_{i}e^{-Y^{T}P^{K}} = \sum_{i}e^{-Y^{T}(P^{K-1}+\alpha_{k}p^{K})}
\]
where $Y = (y_{1}, ..., y_{C})^{T}$ and $i$ denotes the index of the training example. 

%
\vspace{-4mm}
\paragraph{Reweighting Principle}
The key idea of boosting algorithm is to have the current classifier pay more attention to the misclassified samples by previous classifiers. Reweighting is the most common way of budgeting attention based on the historical results. There are essentially two ways to accomplish this goal:
\begin{itemize}[leftmargin=0.5cm]
    \item Reweighting on sampling probabilities: Suppose initially each training example $i$ is assigned $\pi = u_{i} = 1/M$ uniformly, so each sample gets equal chance to be picked. After each round, we reweight the sampling probability according to the classification confidence.
    \item Reweighting on loss/gradient: We may also incorporate $u_{i}$ into the gradient, so that a BNN $b^{k}$ updates parameters with larger step size on misclassified examples and vice versa. For example, set $\nabla_{w}J(b^{k}) \leftarrow \lambda \cdot  (\alpha_{k}p^{k}_{y})(u_{i}) \cdot \nabla_{w}J(b^{k})$, where $\lambda$ is the learning rate. However, we observe that this approach is less effective experimentally for BNNs, and we conjecture that it exaggerates the gradient mismatch problem.
\end{itemize}

\subsection{Test-Time Complexity}

A 1-bit BNN with the same connectivity as the original full-precision 32-bit DNN can save $\sim 32$x memory. In reality, BNN can achieve $\sim 58$x speed up on the current generation of 64-bit CPUs \cite{rastegari2016xnor} and may be further improved with special hardware such as FPGA. Some existing works only binarize the weights but leave activations full-precision, which practically only results in $\gtrsim$ 2x speed up. As for BENN with $K$ ensembles, each BNN's inference is independent, thus the total memory saving is $\sim 32/K$x. As for boosting, we can further compress BNN to save more computations and memory usage. Besides, existing approaches have complexity $O(K^{2})$ with $K$-bit QNN \cite{zhou2016dorefa} or use $K$ binary bases \cite{lin2017towards}, because they cannot avoid the bit collection operation to generate a number, although their fixed-point computation is much more efficient than float-point computation. If $O(1)$ is the time complexity of the boolean operation, then BENN reduces the quadratic complexity to linear, i.e., $O(K)$ with $K$ ensembles but still maintains the very satisfying accuracy and stability as stated above. We can even make the inference in $O(1)$ for BENN if multiple threads are supported. A complete comparison is shown in Table~\ref{table:complex}.

\subsection{Stability Analysis}
\label{sec:analysis}
Given a full-precision real valued DNN $f_{w}$ with a set of parameters $w \sim N(0,\sigma_{w}^{2})$, a BNN $f_{w_{b}}$ with binarized parameters $w_{b}$, input vector $x \sim N(0,1)$ (after Batch Normalization) and perturbation $\Delta x \sim N(0,\sigma^{2})$, and a BENN $f_{w_{\text{benn}}}$ with $K$ ensembles, we want to compare their stability and robustness w.r.t. the network parameters and input perturbation. Here we analyze the variance of output change before and after perturbation, which echoes Eq.~\ref{eq:1} in Sec.~\ref{sec:inspiration}. This is because the output change has zero mean and its variance reflects the distribution of output variation. More specifically, larger variance means increased variation of output w.r.t. input perturbation.

Assume $f_{w}, f_{w_{b}}, f_{w_{\text{benn}}}$ are outputs before non-linear activation function of a single neuron in an one-layer network, we have the output variation of real-value DNN as $f_{w}(x+\Delta x) - f_{w}(x) = w \odot \Delta x$, 
whose distribution has variance $\sigma_{r}^{2} = |w|\sigma_{w}^{2}\sigma^{2}$, where $|w|$ denotes number of input connections for this neuron and $\odot$ denotes inner product. Some modern non-linear activation function $g(\cdot)$ like ReLU will not change the inequality of variances, thus we can omit them in the analysis to keep it simple.

For BNN with both weights and activations binarized, we can rewrite the above formulation as $f_{w_{b}}^{b}(x+\Delta x) - f_{w_{b}}^{b}(x) = \text{sign}(w) \odot [\text{sign}(x+\Delta x) - \text{sign}(x)]$, thus having variance $\sigma_{bnn}^{2} = |w|\sigma_{\text{sign}(w)}^{2}(\sigma_{\text{sign}(x+\Delta x)-\text{sign}(x)}^{2})$. And for BENN-Bagging, we have $\sigma_{benn}^{2} = \sigma_{bnn}^{2} / K$ with $K$ ensembles, since bagging effectively reduces variance. For BENN-Boosting, our model can reduce both bias and variance at the same time. However for boosting, the analysis on bias and variance becomes much more difficult and there are still some debates in literature \cite{buhlmann2007boosting, friedman2000additive}. With these Gaussian assumptions and some numerical experiments (detailed analysis and theorems can be found in supplementary material), we can verify the large stability gain of BENN over BNN compared with floating-number DNN. As for robustness, the same analysis principle can be applied to perturbing weights as $\Delta w$ compared with $\Delta x$ used in stability analysis.

\begin{center}
    \begin{table*}
        \vspace{-3mm}
        \caption{{\small Analysis of Theoretically Computational Complexity on a Single Network. (F-full-precision, $Q_k$-k-bit quantization, B-binary)}}
        \centering
        \scriptsize
        \begin{tabular}{lccccc}
            \toprule
            \toprule
            Network & Weights & Activation & Operations & Memory Saving & Computation Saving  	\\
            \midrule
            Standard DNN &F & F & +, -, $\times$ & 1 & 1   \\
            \hline
            \cite{courbariaux2015binaryconnect,hwang2014fixed,li2016ternary,zhu2016trained,zhou2017incremental},...&B & F & +, - & $\sim$ 32x & $\sim$ 2x   \\
            \hline
            \cite{zhou2016dorefa,hubara2016quantized,wu2016quantized,anwar2015fixed},... &$Q_k$ & $Q_k$ & +, -, $\times$ & $\sim$ $\frac{32}{k}$x & $<$ $\frac{58}{k^2}$x   \\
            \hline
            \cite{lin2017towards},... &$k\times B$ & $k\times B$ & +, -, XNOR, bitcount & $\sim$ $\frac{32}{k}$x & $\sim$ $\frac{58}{k^2}$x  \\
            \hline
            \cite{rastegari2016xnor} and ours&B & B & XNOR, bitcount & $\sim$ 32x & $\sim$ 58x   \\
            \hline
            \bottomrule
        \end{tabular}
        \label{table:complex}
        \vspace{-1mm}
    \end{table*}
\end{center}

\vspace{-12mm}

\section{Independent and Warm-Restart Training for BENNs}
\label{sec:training}

We train our BENN with two different methods. The first one is to initialize each new classifier \emph{independently} and retrain it, which is a traditional way. To accelerate the training of new weak classifier in BENN, we can also initialize the weights of the new classifier by cloning the weights from the most recently trained classifier. We name this training scheme as \emph{warm-restart training}, and we conjecture that the knowledge of those unseen data for the new classifier has been transferred from the inherited weights and is helpful to increase the discriminability of the new classifier.  
Interestingly, we observe that for small network and dataset like Network-In-Network \cite{lin2013network} on CIFAR-10, warm-restart training has better accuracy. However, independent training is better when BENN is applied to large network and dataset such as AlexNet \cite{krizhevsky2012imagenet} and ResNet \cite{he2016deep} on ImageNet since overfitting problem emerges. More discussion can be found in Sec.~\ref{sec:experiments} and Sec.~\ref{sec:discussion}.
\vspace{-4mm}
\paragraph{Implementation Details} We train BENN on the image classification task with CNN block structure containing a batch normalization layer, a binary activation layer, a binary convolution layer, a non-binary activation layer (e.g., sigmoid, ReLU), and a pooling layer, as used by many recent works \cite{rastegari2016xnor, zhou2016dorefa}. To compute the gradient of step function $sign(x)$, we use the same approach suggested by STE. When updating parameters, we use real-valued weights as \cite{rastegari2016xnor} suggests otherwise the tiny update could be killed by deterministic binarization and training cannot move on. In this work, we train each BNN using standard independent and warm-restart training. Unlike the previous works which always keep the first and last layer full-precision, we test 7 different BNN architecture configurations as shown in Table~\ref{table:config} and use them as ingredients for ensemble in BENN.
\begin{table*}
    \caption{Weak BNN Configurations Used to Ensemble (W-weights, A-activation, Params-number of parameters in network). The Last Two are Naive Compressed Network.}
    \centering
    \scriptsize
    \begin{tabular}{lccccccr}
        \toprule
        \toprule
        Weak BNN Configuration/Type (\textit{T}) & Weight & Activation & Size & Params\\
        \midrule
        \multirow{1}{4.5cm}{\textbf{SB}~(Semi-BNN)} & First and last layer:32-bit &  First and last layer:32-bit & 100\% & 100\%\\
        \hline
        \multirow{1}{4.5cm}{\textbf{AB}~(All-BNN)} & All layers:1-bit & All layers:1-bit & 100\% & 100\%\\
        \hline
        \multirow{1}{4.5cm}{\textbf{WQB}~(Weight-Quantized-BNN)} & All layers:Q-bit & All layers:1-bit & 100\% & 100\%\\
        \hline
        \multirow{1}{4.5cm}{\textbf{AQB}~(Activation-Quantized-BNN)} & All layers:1-bit & All layers:Q-bit & 100\% & 100\%\\
        \hline
        \multirow{1}{4.5cm}{\textbf{IB}~(Except-Input-BNN)} & All layers:1-bit & First layer: 32-bit & 100\% & 100\%\\
        \hline
        \multirow{1}{4.5cm}{\textbf{SB/AB/IB-Tiny}~(Tiny-Compress-BNN)} & - & - & 50\% & 25\%\\
        \hline
        \multirow{1}{4.5cm}{\textbf{{}SB/AB/IB-Nano}~(Nano-Compress-BNN)} & - & - & 10\% & 1\%\\
        \hline

        \hline
        \bottomrule
    \end{tabular}
    \label{table:config}
\end{table*}
\vspace{-5mm}
\section{Experimental Results}
\label{sec:experiments}
We evaluate BENN on CIFAR-10 and ImageNet datasets with a self-designed compact Network-In-Network (NIN) \cite{lin2013network}, the standard AlexNet \cite{krizhevsky2012imagenet} and ResNet-18 \cite{he2016deep}, respectively. We have summarized in Table~\ref{table:config} the configurations of all BNN variants. More detailed specifications of the networks can be found in the supplementary material. For each type of BNN, we obtain the converged single BNN (e.g., \textit{SB}) when training is done. We also store BNN after each training step and obtain the best BNN along the way by picking the one with the highest test accuracy (e.g., \textit{Best SB}). We use \textit{BENN-T-R} to denote the \textit{BENN} by aggregating \textit{R} BNNs of configuration \textit{T} (e.g., \textit{BENN-SB-32}). We also denote \textit{Bag/Boost-Indep} and \textit{Bag/Boost-Seq} as bagging/boosting with standard independent training and warm-restart sequential training (Sec.~\ref{sec:training}). All ensembled BNNs share the same network architecture as their real-valued DNN counterpart in this paper, although studying multi-model ensemble is an interesting future work. The code of all our experiments will be made public online.

\begin{table}
    \vspace{-3mm}
        \caption{Oscillation During Training (Instability)}
        \centering
        \scriptsize
        \begin{tabular}{lccccccr}
            \toprule
            \toprule
            Network & Ensemble Method & $\#$Ensemble & STD\\
            \midrule
            \multirow{1}{1.2cm}{SB} & - & 1 & 2.94\\
            \hline
            \multirow{1}{1.2cm}{Best SB}  & - & 1 & 1.40 \\
            \hline
            \multirow{1}{1.2cm}{BENN-SB} & Bag-Seq & 5 & 0.31 \\
            \hline
            \multirow{1}{1.2cm}{BENN-SB} & Boost-Seq & 5 & 0.24 \\
            \hline
            \multirow{1}{1.2cm}{BENN-SB} & Bag-Seq & 32 & 0.03 \\
            \hline
            \multirow{1}{1.2cm}{BENN-SB} & Boost-Seq & 32 & 0.02 \\
            \hline

            \hline
            \bottomrule
        \end{tabular}
        \label{table:osci_tab}
        \vspace{-5mm}
\end{table}

\subsection{Insights Generated from CIFAR-10}
In this section, we show the large performance gain using BENN on CIFAR-10 and summarize some insights. Each BNN is initialized by a pre-trained model from XNOR-Net \cite{rastegari2016xnor} and then retrained by 100 epochs to reach convergence before ensemble. Each full-precision DNN counterpart is trained by 300 epochs to obtain the best accuracy for reference. The learning rate is set to 0.001 and ADAM optimizer is used. Here, we use a compact Network-In-Network (NIN) for CIFAR-10. We first present some significant independent comparisons as follows and then summarize the insights we found.
\begin{figure*}[t]
    \centering
    \includegraphics[width=0.9\linewidth]{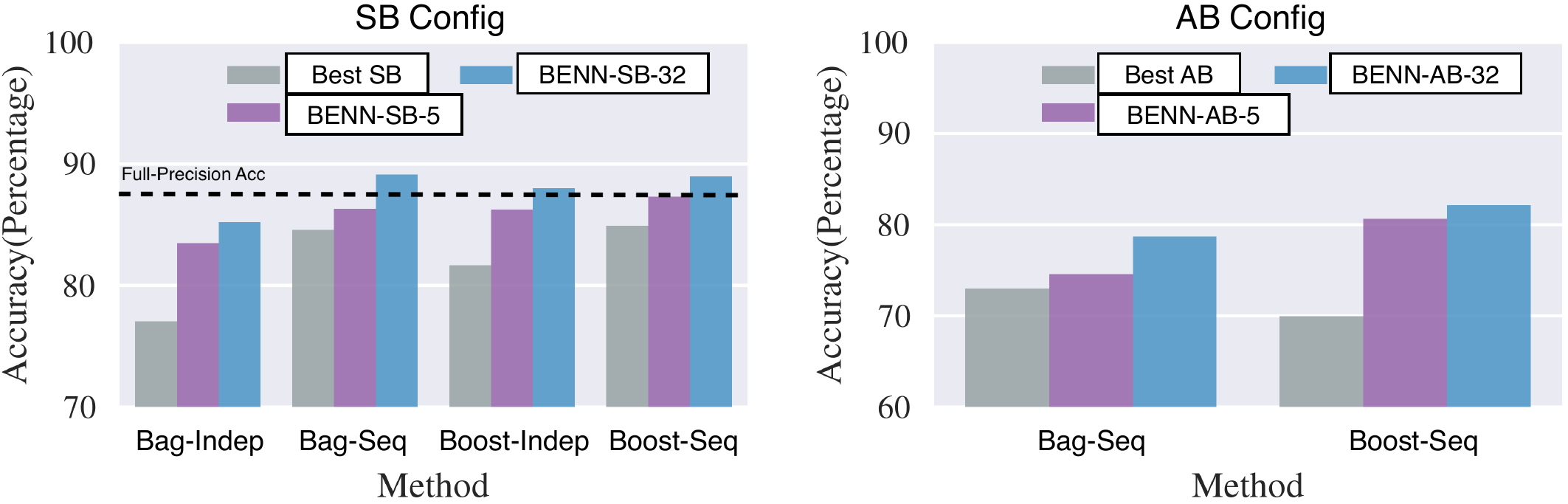}
    \caption{\textbf{Left: }BENN can increase the test accuracy significantly with more ensembles. It can even achieve better accuracy than its full-precision counterpart under Semi-BNN (SB) case. \textbf{Right: }Boosting strongly outperforms bagging in All-BNN (AB) case where each BNN has larger bias.}
    \label{fig:bar1}
    \vspace{-5mm}
\end{figure*}

\textbf{Single BNN versus BENN: } We found that BENN can achieve much better accuracy and stability than a single BNN with negligible sacrifice in speed. Experiments across all BNN configurations show that BENN has the accuracy gain ranging from $4.21\%$ to $24.16\%$ over BNN on CIFAR-10. If each BNN is weak (e.g., AB), the gain of BENN will increase as shown in Fig.~\ref{fig:bar1} (right). This verifies that BNN is indeed a good weak classifier for ensembling. Surprisingly, BENN-SB outperforms full-precision DNN after 32 ensembles (either bagging or boosting) by up to $1.52\%$ (Fig.~\ref{fig:bar1} (left)). Note that in order to have the same memory usage as a 32-bit DNN, we constrain the ensemble up to 32 rounds if no network compression is involved. If more ensembles are available, we observe further performance boost but accuracy gain will eventually become flat.

\begin{figure}
    \vspace{2mm}
    \centering
    \includegraphics[width=\linewidth]{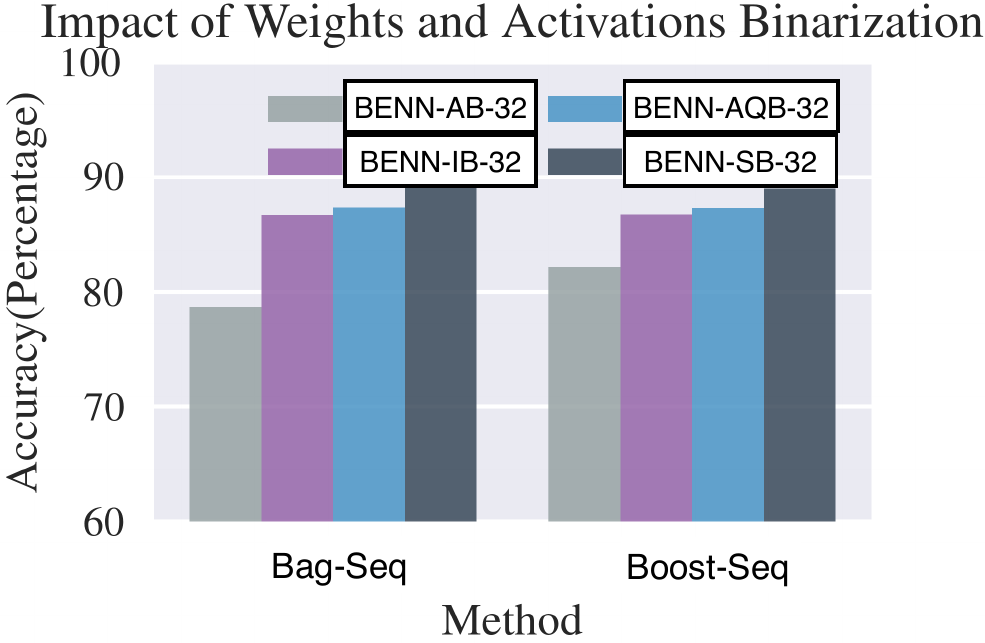}
    \vspace{-3mm}
    \caption{After ensemble, the accuracy increases with more activation bits (Q=2 in AQB). Preserving the first and/or last layer full-precision (IB and SB) helps, compared with all-binary case (AB).}
    \label{fig:bar2}
    \vspace{-5mm}
\end{figure}

We also compare BENN-SB-5 (i.e., 5 ensembles) with WQB (Q=5, 5-bit weight and 1-bit activation), which have the same amount of parameters (measured by bits). WQB can only achieve $\sim 80\%$ accuracy unstably while our ensemble network can reach up to $\sim 86\%$ and remain stable.

We also measure the accuracy variation of the classifier in the last 20 training steps for all BNN configurations. The results in Fig.~\ref{table:osci_tab} indicate that BENN can reduce BNN's variance by $\sim 90\%$ if ensemble 5 rounds and $\sim 99\%$ after 32 rounds. Moreover, picking the best BNN with the highest test accuracy instead of using the BNN when training is done can also reduce the oscillation. This is because the statistical property of ensemble framework (Sec.~\ref{sec:inspiration} and Sec.~\ref{sec:analysis}) makes BENN become a graceful way to ensure high stability.

\textbf{Bagging versus boosting: }It is known that bagging can only reduce the variance of the predictor, while boosting can reduce both bias and variance. Fig.~\ref{fig:bar1}(right), Fig.~\ref{fig:bar2}, and Table~\ref{table:compression} show that boosting outperforms bagging, especially after BNN is compressed, by up to $2.51\%$ when network size is reduced to $50\%$ (\textit{Tiny config}) and $13.38\%$ when network size is reduced to $10\%$ (\textit{Nano config}), and the gain increases from 5 to 32 ensembles. This verifies that boosting is a better choice if the model does not overfit much.

\textbf{Standard independent training versus warm-restart training: }Standard ensemble techniques use independent training, while warm-restart training enable new classifiers to learn faster. Fig.~\ref{fig:bar1}(left) shows that warm-restart training performs better up to $3.9\%$ for bagging and $2.95\%$ for boosting after the same number of training epochs. This means gradually adapting to more examples might be a better choice for CIFAR-10. However, this does not hold for ImageNet task because of slight over-fitting with warm-restart (Sec.~\ref{sec:imagenet}). We believe that this is an interesting phenomenon but it needs more justification by studying the theory of convergence.

\begin{table}
    \vspace{-1mm}
    \caption{Impact of Network Compression}
    \centering
    \scriptsize
    \begin{tabular}{lccccc}
        \toprule
        \toprule
        Network & Ensemble Method & $\#$Ensemble & Accuracy\\
        \midrule
        Best SB & - & 1 & 84.91\%\\
        BENN-SB & Bag-Seq & 32 & 89.12\%\\
        BENN-SB & Boost-Seq & 32 & 89.00\%\\
        \hline
        Best SB-Tiny & - & 1 & 77.20\%\\
        BENN-SB-Tiny & Bag-Seq & 32 & 84.09\%\\
        BENN-SB-Tiny & Boost-Seq & 32 & 84.32\%\\
        \hline
        Best SB-Nano & - & 1 & 40.70\%\\
        BENN-SB-Nano & Bag-Seq & 500 & 57.12\%\\
        BENN-SB-Nano & Boost-Seq & 500 & 63.11\%\\
        \hline
        \bottomrule
    \end{tabular}
    \label{table:compression}
    \vspace{-3mm}
\end{table}

\textbf{The impact of compressing BNN: } BNN's model complexity largely affects bias and variance. If each weak BNN has enough complexity with low bias but high variance, then bagging is more favorable than boosting due to simplicity. However, if each BNN's size is small with large bias, boosting becomes a much better choice. To verify this, we compress each BNN in Table~\ref{table:config} by naively reducing the amount of channels and neurons in each layer. The results in Table~\ref{table:compression} show that BENN-SB can maintain reasonable performance even after naive compression, and boosting gains more over bagging in severe compression (\textit{Nano config}). 

We also found that BENN is less sensitive to network size. Table~\ref{table:compression} shows that compression reduces single BNN's accuracy by $7.71\%$ (\textit{Tiny config}) and $44.21\%$ (\textit{Nano config}). After 32 ensembles, the performance loss caused by compression decreases to $4.8\%$ and $26.01\%$ respectively. Surprisingly, we observe that compression only reduces the accuracy of full-precision DNN by $1.18\%$ (\textit{Tiny config}) and $16.03\%$ (\textit{Nano config}). So it is necessary to have not-too-weak BNNs to build BENN that can compete with full-precision DNN. Better pruning algorithm can be combined with BENN in the future rather than naive compression to allow smaller network to be ensembled.

\textbf{The effect of bit width: } Higher bitwidth results in lower variance and bias at the same time. This can be seen in Fig.~\ref{fig:bar2} where we make activations 2-bit in BENN-AQB (Q=2). As can be seen, BENN-AQB (Q=2) and BENN-IB have comparable accuracy after 32 ensembles, but much better than BENN-AB and worse than BENN-SB. We also observe that activation binarization results in much more unstable model than weight binarization. This indicates that the gain of having more bits is mostly due to better features from the input image, since input binarization is a real pain for neural networks. Surprisingly, BENN-AB can still achieve more than $80\%$ accuracy under such a pain.

\textbf{The effect of binarizing first and last layer: } Almost all the existing works in BNN assume the full precision of the first and last layer, since binarization on these two layers will cause severe accuracy degradation. But we found BENN is less affected, as shown by BENN-AB, BENN-SB and BENN-IB in Fig.~\ref{fig:bar2}. The BNN's accuracy loss due to binarizing these two special layers is $3.98\% \sim 11.9\%$. For BENN with 32 ensembles, the loss reduces to $2.36\% \sim 6.98\%$.

In summary, we generate our main insights about BNN and BENN: (1) Ensemble such as bagging and boosting greatly relieve BNN's problems in terms of representation power, stability, and robustness. (2) Boosting gains advantage over bagging in most cases, and warm-restart training is often a better choice. (3) Weak BNN's configuration (i.e., size, bitwidth, first and last layer) is essential to build a well-functioning BENN to match full-precision DNN in practice.

\subsection{Exploration on Applying BENN to ImageNet Recognition}
\label{sec:imagenet}
We believe BENN is one of the best neural network structures for inference acceleration. To demonstrate the effectiveness of BENN, we compare our algorithm with state-of-the-arts on the ImageNet recognition task (ILSVRC2012) using AlexNet \cite{krizhevsky2012imagenet} and ResNet-18 \cite{he2016deep}. Specifically, we compare our BENN-SB independent training (Sec.~\ref{sec:training}) with the full-precision DNN \cite{krizhevsky2012imagenet, rastegari2016xnor}, DoReFa-Net (k-bit quantized weight and activation) \cite{zhou2016dorefa}, XNOR-Net (binary weight and activation) \cite{rastegari2016xnor}, BNN (binary weight and activation) \cite{hubara2016binarized} and BinaryConnect (binary weight) \cite{courbariaux2015binaryconnect}. We also tried ABC-Net (k binary bases for weight and activation) \cite{lin2017towards} but unfortunately the network does not converge well. Note that accuracy of BNN and BinaryConnect on AlexNet are reported by \cite{rastegari2016xnor} instead of original authors. For DoReFa-Net and ABC-Net, we use the best reported accuracy by original authors with 1-bit weight and 1-bit activation. For XNOR-Net, we report the number of our own retrained model. Our BENN is retrained given a well pre-trained model until convergence by XNOR-Net after 100 epochs to use, and we retrain each BNN with 80 epochs before ensemble. As shown in Table~\ref{table:imgnet} and~\ref{table:resnet}, BENN-SB is the best among all the state-of-the-art BNN architecture, even with only 3 ensembles paralleled on 3 threads. Meanwhile, although we do observe continuous gain with 5 and 8 ensembles (e.g., $56\%$+ on AlexNet), we found that BENN with more ensembles on ImageNet task can be unstable in terms of accuracy and needs further investigation on overfitting issue, otherwise the rapid gain is not always guaranteed. However, we believe our intitial exploration along this direction has shown BENN's potentiality of catching up full-precision DNN and even surpass it with more base BNN classifiers. In fact, how to optimize BENN on large and diverse dataset is still an interesting open problem. 

\vspace{-2mm}
\begin{table}
    \vspace{-3mm}
        \caption{Comparison with state-of-the-arts on ImageNet using AlexNet (W-weights, A-activation)}
        \centering
        \scriptsize
        \begin{tabular}{lccccccr}
            \toprule
            \toprule
            Method & W & A & Top-1\\
            \midrule
            \multirow{1}{3.5cm}{Full-Precision DNN \cite{krizhevsky2012imagenet, rastegari2016xnor}} & 32 & 32 &  56.6\% \\
            \hline
            \multirow{1}{3.5cm}{XNOR-Net \cite{rastegari2016xnor}}  & 1 & 1 &  44.0\%\\
            \hline
            \multirow{1}{3.5cm}{DoReFa-Net \cite{zhou2016dorefa}} & 1 & 1 &  43.6\%\\
            \hline
            \multirow{1}{3.5cm}{BinaryConnect \cite{courbariaux2015binaryconnect, rastegari2016xnor}} & 1 & 32 &  35.4\% \\
            \hline
            \multirow{1}{3.5cm}{BNN \cite{hubara2016binarized, rastegari2016xnor}} & 1 & 1 &  27.9\% \\
            \hline
            \multirow{1}{3.5cm}{\textbf{BENN-SB-3, Bagging (ours)}} & 1 & 1 & \textbf{48.8\%} \\
            \hline
            \multirow{1}{3.5cm}{\textbf{BENN-SB-3, Boosting (ours)}} & 1 & 1 & \textbf{50.2\%} \\
            \hline
            \multirow{1}{3.5cm}{\textbf{BENN-SB-6, Bagging (ours)}} & 1 & 1 & \textbf{52.0\%} \\
            \hline
            \multirow{1}{3.5cm}{\textbf{BENN-SB-6, Boosting (ours)}} & 1 & 1 & \textbf{54.3\%} \\
            \hline
            \bottomrule
        \end{tabular}
        \label{table:imgnet}
\end{table}
\begin{table}
    \vspace{-3mm}
        \caption{Comparison with state-of-the-arts on ImageNet using ResNet-18 (W-weights, A-activation)}
        \centering
        \scriptsize
        \begin{tabular}{lccccccr}
            \toprule
            \toprule
            Method & W & A & Top-1\\
            \midrule
            \multirow{1}{3.5cm}{Full-Precision DNN \cite{he2016deep, lin2017towards}} & 32 & 32 &  69.3\% \\
            \hline
            \multirow{1}{3.5cm}{XNOR-Net \cite{rastegari2016xnor}}  & 1 & 1 & 48.6\%\\
            \hline
            \multirow{1}{3.5cm}{ABC-Net \cite{lin2017towards}}  & 1 & 1 & 42.7\%\\
            \hline
            \multirow{1}{3.5cm}{BNN \cite{hubara2016binarized, rastegari2016xnor}} & 1 & 1 & 42.2\% \\
            \hline
            \multirow{1}{3.5cm}{\textbf{BENN-SB-3, Bagging (ours)}} & 1 & 1 & \textbf{53.4\%} \\
            \hline
            \multirow{1}{3.5cm}{\textbf{BENN-SB-3, Boosting (ours)}} & 1 & 1 & \textbf{53.6\%} \\
            \hline
            \multirow{1}{3.5cm}{\textbf{BENN-SB-6, Bagging (ours)}} & 1 & 1 & \textbf{57.9\%} \\
            \hline
            \multirow{1}{3.5cm}{\textbf{BENN-SB-6, Boosting (ours)}} & 1 & 1 & \textbf{61.0\%} \\
            \hline
            \bottomrule
        \end{tabular}
        \label{table:resnet}
        \vspace{-5mm}
\end{table}

\section{Discussion}
\label{sec:discussion}
\textbf{More bits per network or more networks per bit? } We believe this paper brings up this important question. As for biological neural networks such as our brain, the signal between two neurons is more like a spike instead of high-range real-value signal. This implies that it may not be necessary to use real-valued numbers, while involve a lot of redundancies and can waste significant computing power. Our work converts the direction of ``how many bits per network'' into ``how many networks per bit''. BENN provides a hierarchical view, i.e., we build weak classifiers by  groups of neurons, and build a strong classifier by ensembling the weak classifiers. We have shown that this hierarchical approach is more intuitive and natural to represent knowledge. Although the optimal ensemble structure is beyond the scope of this paper, we believe that some structure searching or meta-learning techniques can be applied. Moreover, the improvement on single BNN such as studying the error surface and resolving the curse of activation/gradient binarization is still essential for the success of BENN.

\textbf{BENN is hardware friendly:} Using BENN with $K$ ensembles is better than using one $K$-bit classifier. Firstly, $K$-bit quantization still cannot get rid of fixed-point multiplication, while BENN can support bitwise operation. People have found that BNN can be further accelerated on FPGAs over modern CPUs \cite{zhao2017accelerating, fu2018towards}. Secondly, people have shown that the complexity of a multiplier is proportional to the square of bitwidth, thus BENN simplifies the hardware design. Thirdly, BENN can use spike signals in the chips instead of keeping the signal real-valued all the time, which can save a lot of energy. Finally, unlike recent literature requiring quadratic time to compute, BENN can be better paralleled on the chips due to its linear time complexity.

\textbf{Current limitations: } It is known to all that ensemble methods can potentially cause overfitting to the model and we also observed similar problems on CIFAR-10 and ImageNet, when the number of ensembles keeps increasing. An interesting next step is to analyze the property of decision boundary of BENN on different datasets and track its evolution in high-dimensional feature space. Also, training will take longer time if many ensembles are needed (especially on large dataset like ImageNet), thus reducing the speed of design iterations. Finally, BENN needs to be further optimized for large networks such as AlexNet and ResNet in order to show its full power, such as picking the best ensemble rule and base classifier.

\section{Conclusion and Future Work}

In this paper, we proposed BENN, a novel neural network architecture which marries BNN with ensemble methods. The experiments showed a large performance gain in terms of accuracy, robustness, and stability. Our experiments also reveal some insights about trade-offs on bit width, network size, number of ensembles, etc. We believe that by leveraging specialized hardware such as FPGA, BENN can be a new dawn for deploying large DNNs into mobile and embedded systems. This work also indicates that a single BNN's properties are still essential thus people need to work hard on both directions. In the future we will explore the power of BENN to reveal more insights about network bit representation and minimal network architecture (e.g., combine BENN with pruning), BENN and hardware co-optimization, and the statistics of BENN's decision boundary.

\small
\nocite{*}
\bibliographystyle{abbrv}
\bibliography{main}

\begin{thebibliography}{10}

\bibitem{anderson2017high}
A.~G. Anderson and C.~P. Berg.
\newblock The high-dimensional geometry of binary neural networks.
\newblock {\em arXiv preprint arXiv:1705.07199}, 2017.

\bibitem{anwar2015fixed}
S.~Anwar, K.~Hwang, and W.~Sung.
\newblock Fixed point optimization of deep convolutional neural networks for
  object recognition.
\newblock In {\em Acoustics, Speech and Signal Processing (ICASSP), 2015 IEEE
  International Conference on}, pages 1131--1135. IEEE, 2015.

\bibitem{bengio2013estimating}
Y.~Bengio, N.~L{\'e}onard, and A.~Courville.
\newblock Estimating or propagating gradients through stochastic neurons for
  conditional computation.
\newblock {\em arXiv preprint arXiv:1308.3432}, 2013.

\bibitem{bousquet2002stability}
O.~Bousquet and A.~Elisseeff.
\newblock Stability and generalization.
\newblock {\em Journal of machine learning research}, 2(Mar):499--526, 2002.

\bibitem{breiman1996bagging}
L.~Breiman.
\newblock Bagging predictors.
\newblock {\em Machine learning}, 24(2):123--140, 1996.

\bibitem{breiman1996bias}
L.~Breiman.
\newblock Bias, variance, and arcing classifiers.
\newblock 1996.

\bibitem{buhlmann2007boosting}
P.~B{\"u}hlmann and T.~Hothorn.
\newblock Boosting algorithms: Regularization, prediction and model fitting.
\newblock {\em Statistical Science}, pages 477--505, 2007.

\bibitem{cai2017deep}
Z.~Cai, X.~He, J.~Sun, and N.~Vasconcelos.
\newblock Deep learning with low precision by half-wave gaussian quantization.
\newblock {\em arXiv preprint arXiv:1702.00953}, 2017.

\bibitem{carney1999confidence}
J.~G. Carney, P.~Cunningham, and U.~Bhagwan.
\newblock Confidence and prediction intervals for neural network ensembles.
\newblock In {\em Neural Networks, 1999. IJCNN'99. International Joint
  Conference on}, volume~2, pages 1215--1218. IEEE, 1999.

\bibitem{courbariaux2015binaryconnect}
M.~Courbariaux, Y.~Bengio, and J.-P. David.
\newblock Binaryconnect: Training deep neural networks with binary weights
  during propagations.
\newblock In {\em Advances in neural information processing systems}, pages
  3123--3131, 2015.

\bibitem{courbariaux2016binarized}
M.~Courbariaux, I.~Hubara, D.~Soudry, R.~El-Yaniv, and Y.~Bengio.
\newblock Binarized neural networks: Training deep neural networks with weights
  and activations constrained to+ 1 or-1.
\newblock {\em arXiv preprint arXiv:1602.02830}, 2016.

\bibitem{deng2009imagenet}
J.~Deng, W.~Dong, R.~Socher, L.-J. Li, K.~Li, and L.~Fei-Fei.
\newblock Imagenet: A large-scale hierarchical image database.
\newblock In {\em Computer Vision and Pattern Recognition, 2009. CVPR 2009.
  IEEE Conference on}, pages 248--255. IEEE, 2009.

\bibitem{deng2017gated}
L.~Deng, P.~Jiao, J.~Pei, Z.~Wu, and G.~Li.
\newblock Gated xnor networks: Deep neural networks with ternary weights and
  activations under a unified discretization framework.
\newblock {\em arXiv preprint arXiv:1705.09283}, 2017.

\bibitem{domingos2000unified}
P.~Domingos.
\newblock A unified bias-variance decomposition.
\newblock In {\em Proceedings of 17th International Conference on Machine
  Learning}, pages 231--238, 2000.

\bibitem{freund1995desicion}
Y.~Freund and R.~E. Schapire.
\newblock A desicion-theoretic generalization of on-line learning and an
  application to boosting.
\newblock In {\em European conference on computational learning theory}, pages
  23--37. Springer, 1995.

\bibitem{freund1996experiments}
Y.~Freund, R.~E. Schapire, et~al.
\newblock Experiments with a new boosting algorithm.
\newblock In {\em Icml}, volume~96, pages 148--156. Bari, Italy, 1996.

\bibitem{friedman2000additive}
J.~Friedman, T.~Hastie, R.~Tibshirani, et~al.
\newblock Additive logistic regression: a statistical view of boosting (with
  discussion and a rejoinder by the authors).
\newblock {\em The annals of statistics}, 28(2):337--407, 2000.

\bibitem{fu2018towards}
C.~Fu, S.~Zhu, H.~Su, C.-E. Lee, and J.~Zhao.
\newblock Towards fast and energy-efficient binarized neural network inference
  on fpga.
\newblock {\em arXiv preprint arXiv:1810.02068}, 2018.

\bibitem{gong2014compressing}
Y.~Gong, L.~Liu, M.~Yang, and L.~Bourdev.
\newblock Compressing deep convolutional networks using vector quantization.
\newblock {\em arXiv preprint arXiv:1412.6115}, 2014.

\bibitem{govindu2004analysis}
G.~Govindu, L.~Zhuo, S.~Choi, and V.~Prasanna.
\newblock Analysis of high-performance floating-point arithmetic on fpgas.
\newblock In {\em Parallel and Distributed Processing Symposium, 2004.
  Proceedings. 18th International}, page 149. IEEE, 2004.

\bibitem{guo2017network}
Y.~Guo, A.~Yao, H.~Zhao, and Y.~Chen.
\newblock Network sketching: Exploiting binary structure in deep cnns.
\newblock {\em arXiv preprint arXiv:1706.02021}, 2017.

\bibitem{gupta2015deep}
S.~Gupta, A.~Agrawal, K.~Gopalakrishnan, and P.~Narayanan.
\newblock Deep learning with limited numerical precision.
\newblock In {\em International Conference on Machine Learning}, pages
  1737--1746, 2015.

\bibitem{han2015deep}
S.~Han, H.~Mao, and W.~J. Dally.
\newblock Deep compression: Compressing deep neural networks with pruning,
  trained quantization and huffman coding.
\newblock {\em arXiv preprint arXiv:1510.00149}, 2015.

\bibitem{han2016incremental}
S.~Han, Z.~Meng, A.-S. Khan, and Y.~Tong.
\newblock Incremental boosting convolutional neural network for facial action
  unit recognition.
\newblock In {\em Advances in Neural Information Processing Systems}, pages
  109--117, 2016.

\bibitem{han2015learning}
S.~Han, J.~Pool, J.~Tran, and W.~Dally.
\newblock Learning both weights and connections for efficient neural network.
\newblock In {\em Advances in neural information processing systems}, pages
  1135--1143, 2015.

\bibitem{hastie2009multi}
T.~Hastie, S.~Rosset, J.~Zhu, and H.~Zou.
\newblock Multi-class adaboost.
\newblock {\em Statistics and its Interface}, 2(3):349--360, 2009.

\bibitem{he2016deep}
K.~He, X.~Zhang, S.~Ren, and J.~Sun.
\newblock Deep residual learning for image recognition.
\newblock In {\em Proceedings of the IEEE conference on computer vision and
  pattern recognition}, pages 770--778, 2016.

\bibitem{hinton}
G.~Hinton.
\newblock Neural networks for machine learning.
\newblock In {\em Coursera}, 2012.

\bibitem{hornik1989multilayer}
K.~Hornik, M.~Stinchcombe, and H.~White.
\newblock Multilayer feedforward networks are universal approximators.
\newblock {\em Neural networks}, 2(5):359--366, 1989.

\bibitem{huang2017learning}
F.~Huang, J.~Ash, J.~Langford, and R.~Schapire.
\newblock Learning deep resnet blocks sequentially using boosting theory.
\newblock {\em arXiv preprint arXiv:1706.04964}, 2017.

\bibitem{hubara2016binarized}
I.~Hubara, M.~Courbariaux, D.~Soudry, R.~El-Yaniv, and Y.~Bengio.
\newblock Binarized neural networks.
\newblock In {\em Advances in neural information processing systems}, pages
  4107--4115, 2016.

\bibitem{hubara2016quantized}
I.~Hubara, M.~Courbariaux, D.~Soudry, R.~El-Yaniv, and Y.~Bengio.
\newblock Quantized neural networks: Training neural networks with low
  precision weights and activations.
\newblock {\em arXiv preprint arXiv:1609.07061}, 2016.

\bibitem{hwang2014fixed}
K.~Hwang and W.~Sung.
\newblock Fixed-point feedforward deep neural network design using weights+ 1,
  0, and- 1.
\newblock In {\em Signal Processing Systems (SiPS), 2014 IEEE Workshop on},
  pages 1--6. IEEE, 2014.

\bibitem{iandola2016squeezenet}
F.~N. Iandola, S.~Han, M.~W. Moskewicz, K.~Ashraf, W.~J. Dally, and K.~Keutzer.
\newblock Squeezenet: Alexnet-level accuracy with 50x fewer parameters and< 0.5
  mb model size.
\newblock {\em arXiv preprint arXiv:1602.07360}, 2016.

\bibitem{ioffe2015batch}
S.~Ioffe and C.~Szegedy.
\newblock Batch normalization: Accelerating deep network training by reducing
  internal covariate shift.
\newblock {\em arXiv preprint arXiv:1502.03167}, 2015.

\bibitem{kim2016bitwise}
M.~Kim and P.~Smaragdis.
\newblock Bitwise neural networks.
\newblock {\em arXiv preprint arXiv:1601.06071}, 2016.

\bibitem{kingma2014adam}
D.~P. Kingma and J.~Ba.
\newblock Adam: A method for stochastic optimization.
\newblock {\em arXiv preprint arXiv:1412.6980}, 2014.

\bibitem{krizhevsky2012imagenet}
A.~Krizhevsky, I.~Sutskever, and G.~E. Hinton.
\newblock Imagenet classification with deep convolutional neural networks.
\newblock In {\em Advances in neural information processing systems}, pages
  1097--1105, 2012.

\bibitem{li2016ternary}
F.~Li, B.~Zhang, and B.~Liu.
\newblock Ternary weight networks.
\newblock {\em arXiv preprint arXiv:1605.04711}, 2016.

\bibitem{li2017training}
H.~Li, S.~De, Z.~Xu, C.~Studer, H.~Samet, and T.~Goldstein.
\newblock Training quantized nets: A deeper understanding.
\newblock In {\em Advances in Neural Information Processing Systems}, pages
  5813--5823, 2017.

\bibitem{lin2016fixed}
D.~Lin, S.~Talathi, and S.~Annapureddy.
\newblock Fixed point quantization of deep convolutional networks.
\newblock In {\em International Conference on Machine Learning}, pages
  2849--2858, 2016.

\bibitem{lin2013network}
M.~Lin, Q.~Chen, and S.~Yan.
\newblock Network in network.
\newblock {\em arXiv preprint arXiv:1312.4400}, 2013.

\bibitem{lin2017towards}
X.~Lin, C.~Zhao, and W.~Pan.
\newblock Towards accurate binary convolutional neural network.
\newblock In {\em Advances in Neural Information Processing Systems}, pages
  344--352, 2017.

\bibitem{lin2015neural}
Z.~Lin, M.~Courbariaux, R.~Memisevic, and Y.~Bengio.
\newblock Neural networks with few multiplications.
\newblock {\em arXiv preprint arXiv:1510.03009}, 2015.

\bibitem{moghimi2016boosted}
M.~Moghimi, S.~J. Belongie, M.~J. Saberian, J.~Yang, N.~Vasconcelos, and L.-J.
  Li.
\newblock Boosted convolutional neural networks.
\newblock In {\em BMVC}, 2016.

\bibitem{ott2016recurrent}
J.~Ott, Z.~Lin, Y.~Zhang, S.-C. Liu, and Y.~Bengio.
\newblock Recurrent neural networks with limited numerical precision.
\newblock {\em arXiv preprint arXiv:1608.06902}, 2016.

\bibitem{oza2001online}
N.~C. Oza and S.~Russell.
\newblock {\em Online ensemble learning}.
\newblock University of California, Berkeley, 2001.

\bibitem{park2017weighted}
E.~Park, J.~Ahn, and S.~Yoo.
\newblock Weighted-entropy-based quantization for deep neural networks.
\newblock In {\em IEEE Conference on Computer Vision and Pattern Recognition
  (CVPR)}, 2017.

\bibitem{polino2018model}
A.~Polino, R.~Pascanu, and D.~Alistarh.
\newblock Model compression via distillation and quantization.
\newblock {\em arXiv preprint arXiv:1802.05668}, 2018.

\bibitem{rastegari2016xnor}
M.~Rastegari, V.~Ordonez, J.~Redmon, and A.~Farhadi.
\newblock Xnor-net: Imagenet classification using binary convolutional neural
  networks.
\newblock In {\em European Conference on Computer Vision}, pages 525--542.
  Springer, 2016.

\bibitem{schapire2003boosting}
R.~E. Schapire.
\newblock The boosting approach to machine learning: An overview.
\newblock In {\em Nonlinear estimation and classification}, pages 149--171.
  Springer, 2003.

\bibitem{schapire2013explaining}
R.~E. Schapire.
\newblock Explaining adaboost.
\newblock In {\em Empirical inference}, pages 37--52. Springer, 2013.

\bibitem{schapire1999improved}
R.~E. Schapire and Y.~Singer.
\newblock Improved boosting algorithms using confidence-rated predictions.
\newblock {\em Machine learning}, 37(3):297--336, 1999.

\bibitem{soudry2014expectation}
D.~Soudry, I.~Hubara, and R.~Meir.
\newblock Expectation backpropagation: Parameter-free training of multilayer
  neural networks with continuous or discrete weights.
\newblock In {\em Advances in Neural Information Processing Systems}, pages
  963--971, 2014.

\bibitem{srivastava2014dropout}
N.~Srivastava, G.~Hinton, A.~Krizhevsky, I.~Sutskever, and R.~Salakhutdinov.
\newblock Dropout: A simple way to prevent neural networks from overfitting.
\newblock {\em The Journal of Machine Learning Research}, 15(1):1929--1958,
  2014.

\bibitem{sung2015resiliency}
W.~Sung, S.~Shin, and K.~Hwang.
\newblock Resiliency of deep neural networks under quantization.
\newblock {\em arXiv preprint arXiv:1511.06488}, 2015.

\bibitem{szegedy2015going}
C.~Szegedy, W.~Liu, Y.~Jia, P.~Sermanet, S.~Reed, D.~Anguelov, D.~Erhan,
  V.~Vanhoucke, A.~Rabinovich, et~al.
\newblock Going deeper with convolutions.
\newblock Cvpr, 2015.

\bibitem{tang2017train}
W.~Tang, G.~Hua, and L.~Wang.
\newblock How to train a compact binary neural network with high accuracy?
\newblock In {\em AAAI}, pages 2625--2631, 2017.

\bibitem{valiant1984theory}
L.~G. Valiant.
\newblock A theory of the learnable.
\newblock {\em Communications of the ACM}, 27(11):1134--1142, 1984.

\bibitem{wan2013regularization}
L.~Wan, M.~Zeiler, S.~Zhang, Y.~Le~Cun, and R.~Fergus.
\newblock Regularization of neural networks using dropconnect.
\newblock In {\em International Conference on Machine Learning}, pages
  1058--1066, 2013.

\bibitem{wu2016quantized}
J.~Wu, C.~Leng, Y.~Wang, Q.~Hu, and J.~Cheng.
\newblock Quantized convolutional neural networks for mobile devices.
\newblock In {\em Proceedings of the IEEE Conference on Computer Vision and
  Pattern Recognition}, pages 4820--4828, 2016.

\bibitem{xu2012robustness}
H.~Xu and S.~Mannor.
\newblock Robustness and generalization.
\newblock {\em Machine learning}, 86(3):391--423, 2012.

\bibitem{zhao2017accelerating}
R.~Zhao, W.~Song, W.~Zhang, T.~Xing, J.-H. Lin, M.~Srivastava, R.~Gupta, and
  Z.~Zhang.
\newblock Accelerating binarized convolutional neural networks with
  software-programmable fpgas.
\newblock In {\em Proceedings of the 2017 ACM/SIGDA International Symposium on
  Field-Programmable Gate Arrays}, pages 15--24. ACM, 2017.

\bibitem{zhou2017incremental}
A.~Zhou, A.~Yao, Y.~Guo, L.~Xu, and Y.~Chen.
\newblock Incremental network quantization: Towards lossless cnns with
  low-precision weights.
\newblock {\em arXiv preprint arXiv:1702.03044}, 2017.

\bibitem{zhou2016dorefa}
S.~Zhou, Y.~Wu, Z.~Ni, X.~Zhou, H.~Wen, and Y.~Zou.
\newblock Dorefa-net: Training low bitwidth convolutional neural networks with
  low bitwidth gradients.
\newblock {\em arXiv preprint arXiv:1606.06160}, 2016.

\bibitem{zhu2016trained}
C.~Zhu, S.~Han, H.~Mao, and W.~J. Dally.
\newblock Trained ternary quantization.
\newblock {\em arXiv preprint arXiv:1612.01064}, 2016.

\end{thebibliography}

\newpage
In this supplementary part, we provide detailed analysis of BNN, DNN and BENN in Sec.~\ref{sec:1}. The training algorithm of BENN is provided in Sec.~\ref{sec:2}, and Sec.~\ref{sec:3} presents network architectures used in our main paper.

\section{Supplementary Material: Detailed Analysis on DNN, BNN, and BENN}
\label{sec:1}

Given a full-precision real valued DNN $f_{w}$ with a set of parameters $w \sim N(0,\sigma_{w}^{2})$, a BNN $f_{w_{b}}$ with binarized parameters $w_{b}$, input vector $x \sim N(0,1)$ (after Batch Normalization) and perturbation $\Delta x \sim N(0,\sigma^{2})$, and a BENN $f_{w_{\text{benn}}}$ with $K$ ensembles, we want to compare their robustness w.r.t. the input perturbation. Here we analyze the variance of output change before and after perturbation, which echoes Eq.1 in Sec.3 in the main paper. This is because the output change has zero mean and its variance reflects the distribution of output variation. More specifically, larger variance means increased variation of output w.r.t. input perturbation.

Assume $f_{w}, f_{w_{b}}, f_{w_{\text{benn}}}$ are outputs before non-linear activation function of a single neuron in an one-layer network, we have the output variation of real-value DNN:

\[
    f_{w}(x+\Delta x) - f_{w}(x) = w \odot (x+\Delta x) - w \odot x = w \odot \Delta x
\]

whose distribution has variance $\sigma_{r}^{2} = |w|\sigma_{w}^{2}\sigma^{2}$, where $|w|$ denotes number of input connections for this neuron and $\odot$ denotes inner product. This is because summation of multiple independent distributions (due to inner product $\odot$) has variance summed as well. Some modern non-linear activation function $g(\cdot)$ like ReLU will not change the inequality of variances (i.e., if $\sigma_{f_{a}}^{2}(x) > \sigma_{f_{b}}^{2}(x)$, then $\sigma_{g(f_{a}(x))}^{2} > \sigma_{g(f_{b}(x))}^{2}$), thus we can omit them in the analysis to keep it simple.

\subsection{Activation Binarization}
\label{sec:acb}
Suppose $w$ is real valued but only input binarized (denote as $f_{w}^{b}$), the activation binarization (-1 and +1) has threshold 0, then the output variation is:

\[
    f_{w}^{b}(x+\Delta x) - f_{w}^{b}(x) = w \odot \text{sign}(x+\Delta x) - w \odot \text{sign}(x)
\]

whose distribution has variance $\sigma_{A_{b}}^{2} = |w|\sigma_{w}^{2}\sigma_{\text{sign}(x+\Delta x)-\text{sign}(x)}^{2}$. This is because $\text{sign}(x)\in\left\{ -1, +1 \right\}$ so the inner product is just the summation of $|w|$ independent distributions, each having variance $\sigma_{\text{sign}(x+\Delta x)-\text{sign}(x)}^{2}$. Note that $\gamma = \text{sign}(x+\Delta x)-\text{sign}(x)$ only has three possible values, namely, 0, -2 and +2. We compute each of them as follows:
\[
    \begin{split}
        \text{Pr}(\gamma = 2) = &\text{Pr}(x < 0 \text{ AND } x+\Delta x > 0)\\
        &= \text{Pr}(\Delta x > -x | x < 0)\text{Pr}(x < 0)\\
        &= \int_{-\infty}^{0^{-}}[\int_{-x}^{\infty}\frac{1}{\sqrt{2\pi}\sigma}e^{-\frac{(\Delta x)^{2}}{2\sigma^{2}}}d(\Delta x)]\frac{1}{\sqrt{2\pi}}e^{-\frac{x^{2}}{2}}dx
    \end{split}
\]
\[
    \begin{split}
        \text{Pr}(\gamma = -2) = &\text{Pr}(x > 0 \text{ AND } x+\Delta x < 0)\\
        &= \text{Pr}(\Delta x < -x | x > 0)\text{Pr}(x > 0)\\
        &= \int_{0^{+}}^{\infty}[\int_{-\infty}^{-x}\frac{1}{\sqrt{2\pi}\sigma}e^{-\frac{(\Delta x)^{2}}{2\sigma^{2}}}d(\Delta x)]\frac{1}{\sqrt{2\pi}}e^{-\frac{x^{2}}{2}}dx\\
    \end{split}
\]
\[
    \begin{split}
        \text{Pr}(\gamma = 0) = 1 - \text{Pr}(\gamma = \pm2)
    \end{split}
\]
and its variance can be computed by:
\[
    \begin{split}
        \sigma_{A_{b}}^{2} &= |w|\sigma_{w}^{2}\left\{\mathbb E[\gamma^{2}] - \mathbb E^{2}[\gamma]\right\}\\
        &= |w|\sigma_{w}^{2}\left\{\mathbb E[\gamma^{2}]\right\}
    \end{split}
\]
since $\mathbb E[(\text{sign}(x+\Delta x)-\text{sign}(x))] = 0$. Unfortunately this integral is too complicated to be solved by analytical formula, thus we use numerical method to obtain $\text{Pr}(\text{sign}(x+\Delta x)-\text{sign}(x) = \pm 2)$. Therefore, the variance is:
\[
    \sigma_{A_{b}}^{2} = B|w|\sigma_{w}^{2}, \sigma_{r}^{2} = R|w|\sigma_{w}^{2} 
\]
where $B$($=\sigma_{\text{sign}(x+\Delta x)-\text{sign}(x)}^{2}$) and $R$($=\sigma^{2}$) can be found in Table~\ref{table:opt_osc_1}. When $\sigma<1$, robustness of BNN is worse than DNN's. As for BENN-Bagging with $K$ ($K>1$) ensembles, the output change has variance:
\[
    \sigma_{\text{benn}}^{2} = K\sigma_{A_{b}}^{2} \cdot \frac{1}{K^{2}} = \frac{\sigma_{A_{b}}^{2}}{K} = \frac{B}{K}|w|\sigma_{w}^{2} < \sigma_{A_{b}}^{2}
\]
thus BENN-Bagging has better robustness than BNN. If $K > \frac{\sigma_{A_{b}}^{2}}{\sigma_{r}^{2}} = \frac{B}{R}$, then BENN-Bagging can have even better robustness than DNN.
\begin{table}
    \caption{Relation between B, R and $\sigma$}
        \centering
        \scriptsize
        \begin{tabular}{lcr}
            \toprule
            \toprule
            $\sigma$ & B & R  \\
            \midrule
            1.5 & \textbf{1.25} & 2.25 \\
            \hline
            1.0 & 1.0 & 1.0 \\
            \hline
            0.5 & 0.59 & \textbf{0.25} \\
            \hline
            0.1 & 0.13 & \textbf{0.01} \\
            \hline
            0.01 & 0.013 & \textbf{0.0001} \\
            \hline
            0.001 & 0.0013 & \textbf{0.000001} \\
            \hline
            \bottomrule
        \end{tabular}
        \label{table:opt_osc_1}
\end{table}
\subsection{Weight Binarization}
\label{sec:wb}
If we binarize $w$ to $w_{b}$ but keeping the activation real-valued, the output variation follows:
\[
    f_{w_{b}}(x+\Delta x) - f_{w_{b}}(x) = \text{sign}(w) \odot \Delta x
\]
with variance $\sigma_{W_{b}}^{2} =  |w|\sigma_{\text{sign}(w)}^{2}\sigma^{2} = |w|\sigma^{2}$. Thus whether weight binarization will hurt robustness or not depends on whether $\sigma_{\text{sign}(w)}^{2} = 1 > \sigma_{w}^{2}$ holds or not. In particular, the robustness will not decrease if $\sigma_{w}^{2} = 1$. BENN-Bagging has variance $\sigma_{\text{benn}}^{2} = \frac{1}{K}|w|\sigma^{2}$. So if $K > \frac{1}{\sigma_{w}^{2}}$, then BENN-Bagging is better than DNN.

\subsection{Binarization of Both Weight and Activation}
If both activation and weight are binarized (denote as $f_{w_{b}}^{b}$), the output variation:
\[
    f_{w_{b}}^{b}(x+\Delta x) - f_{w_{b}}^{b}(x) = \text{sign}(w) \odot [\text{sign}(x+\Delta x) - \text{sign}(x)]
\]
has variance $\sigma_{E_{b}}^{2} = |w|\sigma_{\text{sign}(w)}^{2}(\sigma_{\text{sign}(x+\Delta x)-\text{sign}(x)}^{2}) = B|w|$ which is just the combination of Sec.~\ref{sec:acb} and Sec.~\ref{sec:wb}. BENN-Bagging has variance $\sigma_{\text{benn}}^{2} = \frac{B}{K}|w|$, which is more robust than DNN when $K > \frac{\sigma_{E_{b}}^{2}}{\sigma_{r}^{2}} = \frac{B}{R\sigma_{w}^{2}}$.

The above analysis results in the following theorem:

\begin{theorem}
    \label{lm1}
    Given a activation binarization, weight binarization or extreme binarization one-layer network introduced above, input perturbation is $\Delta x \sim N(0,\sigma^{2})$, then the output variation obeys:
    \begin{enumerate}
        \item If only activation is binarized, BNN has worse robustness than DNN when perturbation $\sigma < 1$. BENN-Bagging is guaranteed to be more robust than BNN. BENN-Bagging with $K$ ensembles is more robust than DNN when $K > \frac{B}{R}$.
        \item If only weight is binarized, BNN has worse robustness than DNN when $\sigma_{w} < 1$. BENN-Bagging is guaranteed to be more robust than BNN. BENN-Bagging with $K$ ensembles is more robust than DNN when $K > \frac{1}{\sigma_{w}^{2}}$.
        \item If both weight and activation are binarized, BNN has worse robustness than DNN when $\sigma_{w} < 1$ and perturbation $\sigma < 1$. BENN-Bagging is guaranteed to be more robust than BNN. BENN-Bagging with $K$ ensembles is more robust than DNN when $K > \frac{B}{R\sigma_{w}^{2}}$.
    \end{enumerate}

\end{theorem}

\subsection{Multiple Layers Scenario}
All the above analysis is for one layer models before and after activation function. The same conclusion can be extended to multiple layers scenario with Theorem~\ref{lm1}.
\begin{theorem}
    \label{lm1}
    Given a activation binarization, weight binarization or extreme binarization L-layer network (without batch normalization for generalization) introduced above, input perturbation is $\Delta x \sim N(0,\sigma^{2})$, then the accumulated perturbation of ultimate network output obeys:
    \begin{enumerate}
        \item For DNN, ultimate output variation is $\sigma_r^2 \leq \sigma^2\prod_{l=1}^{L}|w_l|\sigma_{w_l}^2$.
        \item For activation binarization BNN, ultimate output variation is $\sigma_{A_b}^2\leq B\prod_{l=1}^{L}|w_l|\sigma_{w_l}^2$.
        \item For weight binarization BNN, ultimate output variation is $\sigma_{W_b}^2 \leq \sigma^2\prod_{l=1}^{L}|w_l|$
        \item For extreme binarization BNN, ultimate output variation is $\sigma_{E_b}^2\leq B \prod_{l=1}^{L} |w_l|$.
        \item Theorem~1 holds for multiple layers scenario.
    \end{enumerate}

\end{theorem}

People have not fully understood the effect of variance reduction in boosting algorithms and some debates still exist in literature \cite{buhlmann2007boosting, friedman2000additive}, given that classifiers are not independent with each other. However, our experiments show that BENN-boosting can also reduce variance in our situation, which is consistent with \cite{freund1996experiments, friedman2000additive}. The theoretical analysis on BENN-boosting is left for future work.

If we switch $x$ and $w$, replace input perturbation $\Delta x$ with parameter perturbation $\Delta w$ in the above analysis, then the same conclusion holds for parameter perturbation (stability issue). To sum up, BNN often can be worse than DNN in terms of robustness and stability, and our method BENN can cure these problems.

\newpage
\section{Supplementary Material: Training Process of BENN}
\label{sec:2}
\begin{algorithm}
	\SetAlgoLined
	\textbf{Input:} a full-precision neural net with $L$ layers, $f$ elements in convolution kernel and learning rate $\eta$, initial weight $u_{i}$ for each training example $i$ and number of ensemble rounds $K$. Initialize BNN with a pre-trained XNOR-Net model \cite{rastegari2016xnor}. Retrain each BNN for maximally $M$ epochs.
	
	\textbf{Ensemble Pass:}
	
	\For{k=1 to K}{
	    Sampling a new training set given weight $u_{i}$ of each example $i$;
	
	    \For{epoch=1 to M}{
	    
	    \textbf{Forward Pass: }
	
	    \For{$l$=1 to $L$}{
	        \For{each filter in $l$-th layer}{
	            $a^{l} = \frac{1}{f}||w_{t}^{l}||_{l1}$;
	        
	            $b^{l} = \text{Sign}(w_{t}^{l})$;
	        
	            $w_{b}^{l} = a^{l}b^{l}$
	        }
	        Compute activation $a_{l}$ based on binary kernel $w_{b}^{l}$ and input $a_{l-1}$;
	    }
	
	    \textbf{Backward Pass: }
	
	    Compute gradient $\frac{\partial J}{\partial w_{t}}$ based on \cite{rastegari2016xnor, hinton};
	
	    \textbf{Parameter Update: }
	
	    Update $w_{t}$ to $w_{t+1}$ with any update rules (e.g., SGD or ADAM)
	    
	    }
	    
	    \textbf{Ensemble Update: }
	    
	    Pick the BNN when training converges;
	    
	    Use either bagging or boosting algorithm to update weight $u_{i}$ of each training example $i$;
	}
	
    \textbf{Return:} $K$ trained base classifiers for BENN;
	
	\caption{Training Process of BENN}
\end{algorithm}

\newpage
\section{Supplementary Material: Network Architectures Used in the Paper}
\label{sec:3}
In this section we provide network architectures used in the experiments of our main paper.

\subsubsection{Self-Designed Network-In-Network (NIN)}
\begin{table}[ht]
    \caption{Self-Designed Network-In-Network (NIN)}
    \centering
    \scriptsize
    \begin{tabular}{|l|c|c|r}
        \toprule
        \toprule
        Layer Index & Type & Parameters \\
        \midrule
        \multirow{1}{0.5cm}{1} & Conv & Depth: 192, Kernel Size: 5x5, Stride: 1, Padding: 2\\
        \multirow{1}{0.5cm}{2} & BatchNorm & $\epsilon$: 0.0001, Momentum: 0.1\\
        \multirow{1}{0.5cm}{3} & ReLU & -\\
        
        \multirow{1}{0.5cm}{4} & BatchNorm & $\epsilon$: 0.0001, Momentum: 0.1\\
        \multirow{1}{0.5cm}{5} & Dropout & $p$: 0.5\\
        \multirow{1}{0.5cm}{6} & Conv & Depth: 96, Kernel Size: 1x1, Stride: 1, Padding: 0\\
        \multirow{1}{0.5cm}{7} & ReLU & -\\
        
        \multirow{1}{0.5cm}{8} & MaxPool & Kernel: 3x3, Stride: 2, Padding: 1\\
        
        \multirow{1}{0.5cm}{9} & BatchNorm & $\epsilon$: 0.0001, Momentum: 0.1\\
        \multirow{1}{0.5cm}{10} & Dropout & $p$: 0.5\\
        \multirow{1}{0.5cm}{11} & Conv & Depth: 192, Kernel Size: 5x5, Stride: 1, Padding: 2\\
        \multirow{1}{0.5cm}{12} & ReLU & -\\
        
        \multirow{1}{0.5cm}{13} & BatchNorm & $\epsilon$: 0.0001, Momentum: 0.1\\
        \multirow{1}{0.5cm}{14} & Dropout & $p$: 0.5\\
        \multirow{1}{0.5cm}{15} & Conv & Depth: 192, Kernel Size: 1x1, Stride: 1, Padding: 0\\
        \multirow{1}{0.5cm}{16} & ReLU & -\\
        
        \multirow{1}{0.5cm}{17} & AvgPool & Kernel: 3x3, Stride: 2, Padding: 1\\
        
        \multirow{1}{0.5cm}{18} & BatchNorm & $\epsilon$: 0.0001, Momentum: 0.1\\
        \multirow{1}{0.5cm}{19} & Dropout & $p$: 0.5\\
        \multirow{1}{0.5cm}{20} & Conv & Depth: 192, Kernel Size: 3x3, Stride: 1, Padding: 1\\
        \multirow{1}{0.5cm}{21} & ReLU & -\\
        
        \multirow{1}{0.5cm}{22} & BatchNorm & $\epsilon$: 0.0001, Momentum: 0.1\\
        \multirow{1}{0.5cm}{23} & Conv & Depth: 192, Kernel Size: 1x1, Stride: 1, Padding: 0\\
        \multirow{1}{0.5cm}{24} & ReLU & -\\
        
        \multirow{1}{0.5cm}{25} & BatchNorm & $\epsilon$: 0.0001, Momentum: 0.1\\
        \multirow{1}{0.5cm}{26} & Conv & Depth: 192, Kernel Size: 1x1, Stride: 1, Padding: 0\\
        \multirow{1}{0.5cm}{27} & ReLU & -\\
        
        \multirow{1}{0.5cm}{28} & AvgPool & Kernel: 8x8, Stride: 1, Padding: 0\\
        \multirow{1}{0.5cm}{29} & FC & Width: 1000\\
        
        \hline

        \hline
        \bottomrule
    \end{tabular}
    \label{table:var_nin}
    \vspace{-2mm}
\end{table}

\subsubsection{AlexNet}
\begin{table}[ht]
    \caption{AlexNet}
    \centering
    \scriptsize
    \begin{tabular}{|l|c|c|r}
        \toprule
        \toprule
        Layer Index & Type & Parameters \\
        \midrule
        \multirow{1}{0.5cm}{1} & Conv & Depth: 96, Kernel Size: 11x11, Stride: 4, Padding: 0\\
        \multirow{1}{0.5cm}{2} & ReLU & -\\
        \multirow{1}{0.5cm}{3} & MaxPool & Kernel: 3x3, Stride: 2\\
        \multirow{1}{0.5cm}{4} & BatchNorm & -\\
        
        \multirow{1}{0.5cm}{5} & Conv & Depth: 256, Kernel Size: 5x5, Stride: 1, Padding: 2\\
        \multirow{1}{0.5cm}{6} & ReLU & -\\
        \multirow{1}{0.5cm}{7} & MaxPool & Kernel: 3x3, Stride: 2\\
        \multirow{1}{0.5cm}{8} & BatchNorm & -\\
        
        \multirow{1}{0.5cm}{9} & Conv & Depth: 384, Kernel Size: 3x3, Stride: 1, Padding: 1\\
        \multirow{1}{0.5cm}{10} & ReLU & -\\
        \multirow{1}{0.5cm}{11} & Conv & Depth: 384, Kernel Size: 3x3, Stride: 1, Padding: 1\\
        \multirow{1}{0.5cm}{12} & ReLU & -\\
        
        \multirow{1}{0.5cm}{13} & Conv & Depth: 256, Kernel Size: 3x3, Stride: 1, Padding: 1\\
        \multirow{1}{0.5cm}{14} & ReLU & -\\
        \multirow{1}{0.5cm}{15} & MaxPool & Kernel: 3x3, Stride: 2\\
        
        \multirow{1}{0.5cm}{16} & Dropout & $p$: 0.5\\
        \multirow{1}{0.5cm}{17} & FC & Width: 4096\\
        \multirow{1}{0.5cm}{18} & ReLU & -\\
        
        \multirow{1}{0.5cm}{19} & Dropout & $p$: 0.5\\
        \multirow{1}{0.5cm}{20} & FC & Width: 4096\\
        \multirow{1}{0.5cm}{21} & ReLU & -\\
        
        \multirow{1}{0.5cm}{22} & FC & Width: 1000\\
        
        \hline

        \hline
        \bottomrule
    \end{tabular}
    \label{table:alex_chart}
    \vspace{-2mm}
\end{table}

\subsubsection{ResNet-18}
\begin{table}[ht]
    \vspace{-5mm}
    \caption{ResNet-18}
    \centering
    \scriptsize
    \begin{tabular}{|l|c|c|r}
        \toprule
        \toprule
        Layer Index & Type & Parameters \\
        \midrule
        \multirow{1}{0.5cm}{1} & Conv & Depth: 64, Kernel Size: 7x7, Stride: 2, Padding: 3\\
        \multirow{1}{0.5cm}{2} & BatchNorm & $\epsilon$: 0.00001, Momentum: 0.1\\
        \multirow{1}{0.5cm}{3} & ReLU & -\\
        \multirow{1}{0.5cm}{4} & MaxPool & Kernel: 3x3, Stride: 2\\
        
        \multirow{1}{0.5cm}{5} & Conv & Depth: 64, Kernel Size: 3x3, Stride: 1, Padding: 1\\
        \multirow{1}{0.5cm}{6} & BatchNorm & $\epsilon$: 0.00001, Momentum: 0.1\\
        \multirow{1}{0.5cm}{7} & ReLU & -\\
        \multirow{1}{0.5cm}{8} & Conv & Depth: 64, Kernel Size: 3x3, Stride: 1, Padding: 1\\
        \multirow{1}{0.5cm}{9} & BatchNorm & $\epsilon$: 0.00001, Momentum: 0.1\\
        \multirow{1}{0.5cm}{10} & Conv & Depth: 64, Kernel Size: 3x3, Stride: 1, Padding: 1\\
        \multirow{1}{0.5cm}{11} & BatchNorm & $\epsilon$: 0.00001, Momentum: 0.1\\
        \multirow{1}{0.5cm}{12} & ReLU & -\\
        \multirow{1}{0.5cm}{13} & Conv & Depth: 64, Kernel Size: 3x3, Stride: 1, Padding: 1\\
        \multirow{1}{0.5cm}{14} & BatchNorm & $\epsilon$: 0.00001, Momentum: 0.1\\

        \multirow{1}{0.5cm}{15} & Conv & Depth: 128, Kernel Size: 3x3, Stride: 2, Padding: 1\\
        \multirow{1}{0.5cm}{16} & BatchNorm & $\epsilon$: 0.00001, Momentum: 0.1\\
        \multirow{1}{0.5cm}{17} & ReLU & -\\
        \multirow{1}{0.5cm}{18} & Conv & Depth: 128, Kernel Size: 3x3, Stride: 1, Padding: 1\\
        \multirow{1}{0.5cm}{19} & BatchNorm & $\epsilon$: 0.00001, Momentum: 0.1\\
        \multirow{1}{0.5cm}{20} & Conv & Depth: 128, Kernel Size: 1x1, Stride: 2\\
        \multirow{1}{0.5cm}{21} & BatchNorm & $\epsilon$: 0.00001, Momentum: 0.1\\
        \multirow{1}{0.5cm}{22} & Conv & Depth: 128, Kernel Size: 3x3, Stride: 1, Padding: 1\\
        \multirow{1}{0.5cm}{23} & BatchNorm & $\epsilon$: 0.00001, Momentum: 0.1\\
        \multirow{1}{0.5cm}{24} & ReLU & -\\
        \multirow{1}{0.5cm}{25} & Conv & Depth: 128, Kernel Size: 3x3, Stride: 1, Padding: 1\\
        \multirow{1}{0.5cm}{26} & BatchNorm & $\epsilon$: 0.00001, Momentum: 0.1\\

        \multirow{1}{0.5cm}{27} & Conv & Depth: 256, Kernel Size: 3x3, Stride: 2, Padding: 1\\
        \multirow{1}{0.5cm}{28} & BatchNorm & $\epsilon$: 0.00001, Momentum: 0.1\\
        \multirow{1}{0.5cm}{29} & ReLU & -\\
        \multirow{1}{0.5cm}{30} & Conv & Depth: 256, Kernel Size: 3x3, Stride: 1, Padding: 1\\
        \multirow{1}{0.5cm}{31} & BatchNorm & $\epsilon$: 0.00001, Momentum: 0.1\\
        \multirow{1}{0.5cm}{32} & Conv & Depth: 256, Kernel Size: 1x1, Stride: 2\\
        \multirow{1}{0.5cm}{33} & BatchNorm & $\epsilon$: 0.00001, Momentum: 0.1\\
        \multirow{1}{0.5cm}{34} & Conv & Depth: 256, Kernel Size: 3x3, Stride: 1, Padding: 1\\
        \multirow{1}{0.5cm}{35} & BatchNorm & $\epsilon$: 0.00001, Momentum: 0.1\\
        \multirow{1}{0.5cm}{36} & ReLU & -\\
        \multirow{1}{0.5cm}{37} & Conv & Depth: 256, Kernel Size: 3x3, Stride: 1, Padding: 1\\
        \multirow{1}{0.5cm}{38} & BatchNorm & $\epsilon$: 0.00001, Momentum: 0.1\\

        \multirow{1}{0.5cm}{39} & Conv & Depth: 512, Kernel Size: 3x3, Stride: 2, Padding: 1\\
        \multirow{1}{0.5cm}{40} & BatchNorm & $\epsilon$: 0.00001, Momentum: 0.1\\
        \multirow{1}{0.5cm}{41} & ReLU & -\\
        \multirow{1}{0.5cm}{42} & Conv & Depth: 512, Kernel Size: 3x3, Stride: 1, Padding: 1\\
        \multirow{1}{0.5cm}{43} & BatchNorm & $\epsilon$: 0.00001, Momentum: 0.1\\
        \multirow{1}{0.5cm}{44} & Conv & Depth: 512, Kernel Size: 1x1, Stride: 2\\
        \multirow{1}{0.5cm}{45} & BatchNorm & $\epsilon$: 0.00001, Momentum: 0.1\\
        \multirow{1}{0.5cm}{46} & Conv & Depth: 512, Kernel Size: 3x3, Stride: 1, Padding: 1\\
        \multirow{1}{0.5cm}{47} & BatchNorm & $\epsilon$: 0.00001, Momentum: 0.1\\
        \multirow{1}{0.5cm}{48} & ReLU & -\\
        \multirow{1}{0.5cm}{49} & Conv & Depth: 512, Kernel Size: 3x3, Stride: 1, Padding: 1\\
        \multirow{1}{0.5cm}{50} & BatchNorm & $\epsilon$: 0.00001, Momentum: 0.1\\
        
        \multirow{1}{0.5cm}{51} & AvgPool & -\\
        \multirow{1}{0.5cm}{52} & FC & Width: 1000\\
        
        \hline
        
        \hline
        \bottomrule
    \end{tabular}
    \label{table:res_chart}
    \vspace{-6mm}
\end{table}

\end{document}